\documentclass[sigconf]{acmart}

\usepackage{algorithm}
\usepackage{algorithmic}
\usepackage{amsmath}
\usepackage{amsthm}
\usepackage{pifont}
\usepackage{mathtools}
\usepackage{booktabs}
\usepackage{boldline} 
\usepackage{multirow}
\usepackage{colortbl}
\usepackage{graphicx}
\usepackage{arydshln}
\usepackage{makecell}
\usepackage{balance}
\AtBeginDocument{%
  }

\copyrightyear{2025}
\acmYear{2025}
\setcopyright{cc}
\setcctype{by}
\acmConference[KDD '25]{Proceedings of the 31st ACM SIGKDD Conference on
Knowledge Discovery and Data Mining V.2}{August 3--7, 2025}{Toronto, ON,
Canada}
\acmBooktitle{Proceedings of the 31st ACM SIGKDD Conference on Knowledge
Discovery and Data Mining V.2 (KDD '25), August 3--7, 2025, Toronto, ON,
Canada}
\acmDOI{10.1145/3711896.3736566}
\acmISBN{979-8-4007-1454-2/2025/08}

\begin{document}

\title[Graph Prompting for Graph Learning Models: Recent Advances and Future Directions]{Graph Prompting for Graph Learning Models: Recent Advances and Future Directions}


\author{Xingbo Fu}
\affiliation{%
  \institution{University of Virginia}
 \city{Charlottesville}
 \state{VA}
 \country{USA}
 }
\email{xf3av@virginia.edu}

\author{Zehong Wang}
\affiliation{%
  \institution{University of Notre Dame}
 \city{Notre Dame}
 \state{IN}
 \country{USA}
 }
\email{zwang43@nd.edu}

\author{Zihan Chen}
\affiliation{%
  \institution{University of Virginia}
 \city{Charlottesville}
 \state{VA}
 \country{USA}
 }
\email{brf3rx@virginia.edu}

\author{Jiazheng Li}
\affiliation{%
  \institution{University of Connecticut}
 \city{Storrs}
 \state{CT}
 \country{USA}
 }
\email{jiazheng.li@uconn.edu}

\author{Yaochen Zhu}
\affiliation{%
  \institution{University of Virginia}
 \city{Charlottesville}
 \state{VA}
 \country{USA}
 }
\email{uqp4qh@virginia.edu}

\author{Zhenyu Lei}
\affiliation{%
  \institution{University of Virginia}
 \city{Charlottesville}
 \state{VA}
 \country{USA}
 }
\email{vjd5zr@virginia.edu}

\author{Cong Shen}
\affiliation{%
  \institution{University of Virginia}
 \city{Charlottesville}
 \state{VA}
 \country{USA}
 }
\email{cong@virginia.edu}

\author{Yanfang Ye}
\affiliation{%
  \institution{University of Notre Dame}
 \city{Notre Dame}
 \state{IN}
 \country{USA}
 }
\email{yye7@nd.edu}

\author{Chuxu Zhang}
\affiliation{%
  \institution{University of Connecticut}
 \city{Storrs}
 \state{CT}
 \country{USA}
 }
\email{chuxu.zhang@uconn.edu}

\author{Jundong Li}
\affiliation{%
  \institution{University of Virginia}
 \city{Charlottesville}
 \state{VA}
 \country{USA}
 }
\email{jundong@virginia.edu}


\renewcommand{\shortauthors}{Xingbo Fu et al.}

\begin{abstract}
Graph learning models have demonstrated great prowess in learning expressive representations from large-scale graph data in a wide variety of real-world scenarios.
As a prevalent strategy for training powerful graph learning models, the "pre-training, adaptation" scheme first pre-trains graph learning models on unlabeled graph data in a self-supervised manner and then adapts them to specific downstream tasks.
During the adaptation phase, graph prompting emerges as a promising approach that learns trainable prompts while keeping the pre-trained graph learning models unchanged.
In this paper, we present a systematic review of recent advancements in graph prompting.
First, we introduce representative graph pre-training methods that serve as the foundation step of graph prompting.
Next, we review mainstream techniques in graph prompting and elaborate on how they design learnable prompts for graph prompting.
Furthermore, we summarize the real-world applications of graph prompting from different domains.
Finally, we discuss several open challenges in existing studies with promising future directions in this field.
\end{abstract}




\maketitle

\section{Introduction} 

Graphs are prevalent in a wide range of real-world scenarios, such as bioinformatics~\cite{yang2020graph}, traffic networks~\cite{qi2022graph}, and healthcare systems~\cite{fu2023spatial}. 
To gain deep insights from massive graph data, numerous graph learning models, such as graph neural networks (GNNs)~\cite{kipf2016gcn,hamilton2017graphsage,velivckovic2017gat,wang2019dynamic,rossi2020tgn} and graph transformers~\cite{yun2019graph, hu2020heterogeneous, rampavsek2022recipe}, have been developed in recent years and have shown great prowess in different graph-related downstream tasks, including node classification~\cite{luan2023graph,zhao2024imbalanced}, link prediction~\cite{hwang2022analysis,zhu2024pitfalls}, and graph classification~\cite{li2022geomgcl,zhuang2024imold}.
Traditionally, graph learning models are trained via supervised learning.
However, the supervised manner relies heavily on sufficient labeled graph data, which may be infeasible in the real world. 
Additionally, the trained graph learning models cannot be well generalized to other downstream tasks, even on
the same graph data. 
These two critical limitations hinder further practical deployments of graph learning models.

To overcome the above limitations, numerous recent studies have investigated designing effective pre-training frameworks for training powerful graph learning models without using any label information from downstream tasks~\citep{velickovic2019deep, hu2019pre, you2020graphcl, xia2022simgrace, hou2022graphmae, hou2023graphmae2,wang2025git,wang2024gft}.
The philosophy behind graph pre-training is to pre-train graph learning models by learning expressive graph representations via self-supervised learning.
Typically, these graph representations embed generalizable knowledge from graph data and can be adopted for downstream tasks.
However, there exist inevitable objective gaps between pre-training and downstream tasks.
For example, a graph learning model can be pre-trained via link prediction and later adapted for node classification as the downstream task.
In this example, we are more interested in the relationship between a pair of nodes for link prediction during pre-training, while node classification as the downstream task focuses more on the representation of every single node.
As a result, graph representations learned by pre-trained graph learning models may be detrimental to downstream tasks.

\begin{figure}[!t] 
\centerline{\includegraphics[width=0.97\columnwidth]{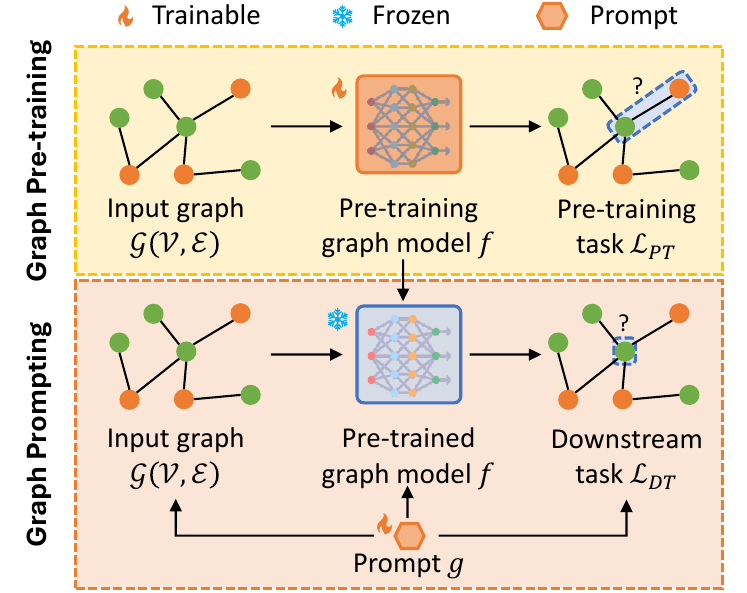}}
\caption{The workflow of the "pre-training, prompting" scheme in graph learning.}
\label{figure:workflow}
\vskip -0.1in
\end{figure}

To bridge the objective gaps, the "pre-training, prompting" scheme has been widely adopted by numerous studies in graph learning~\cite{sun2022gppt, liu2023graphprompt, sun2023all, fang2023gpf, yu2024non, fu2025edge}. 
Inspired by recent prompt tuning approaches in natural language processing~\cite{zhou2022coop,khattak2023maple} and computer vision~\cite{jia2022vpt,yoo2023improving}, graph prompting adapts pre-trained graph learning models by learning additional prompt vectors while keeping the pre-trained graph learning models frozen.
Figure~\ref{figure:workflow} illustrates the workflow of the "pre-training, prompting" scheme.
Compared to fine-tuning~\cite{zhili2024search, huang2024measuring}, graph prompting only needs to update far fewer parameters, making it more suitable for the few-shot setting.
Moreover, graph prompting is more flexible since learnable graph prompts can be designed at different levels for downstream tasks.
These advantages make graph prompting a promising topic in graph learning, encouraging researchers and practitioners to further explore its techniques in both academia and industry.

\begin{figure*}[!t] 
\begin{center}
\centerline{\includegraphics[width=\textwidth]{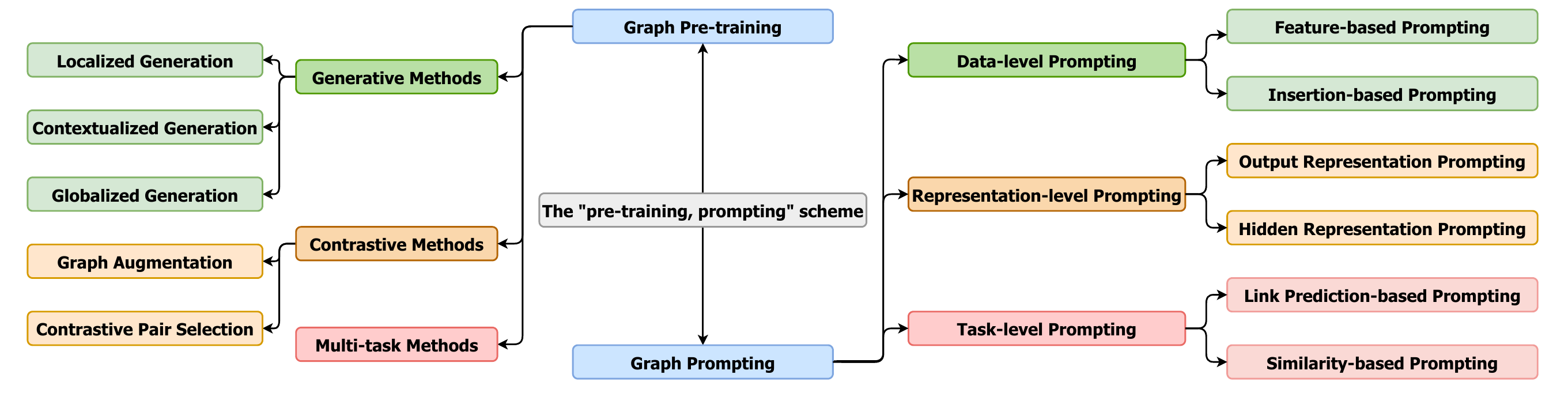}}
\caption{Taxonomy of techniques in the "pre-training, prompting" scheme.}
\label{figure:pretrain}
\end{center}
\end{figure*}

In this survey, we provide a comprehensive and up-to-date introduction to the existing literature in the context of graph prompting.
Specifically, we first introduce representative graph pre-training methods that serve as the foundation stage of graph prompting.
Next, we provide an organized summary of existing techniques in graph prompting.
Compared to previous survey papers~\cite{sun2023survey, long2024towards}, we propose a novel taxonomy of graph prompting methods from the perspectives of graph data, node representations, and downstream tasks.
We further present important applications of graph prompting techniques in various real-world domains.
Finally, we discuss the research challenges and open questions in existing studies, aiming to encourage further advancements in this area.
We summarize the main contributions of our survey paper as follows.
\begin{itemize}
    \item \textbf{Novel taxonomy.} Our survey paper provides a novel taxonomy of existing graph pre-training and prompting methods illustrated in Figure~\ref{figure:pretrain}.

    \item \textbf{Recent techniques.} Compared to previous survey papers on graph prompting, more recent graph prompting techniques are included and introduced in our survey paper.

    \item \textbf{New directions.} Based on existing techniques, we provide insights into future directions for graph prompting from different perspectives.
\end{itemize}

\section{Preliminaries} 

\subsection{Notations and Definitions}
First, we introduce key notations and definitions in graph learning.
For the general purpose, we primarily focus on attributed graphs in this survey paper. 
We use $\mathcal{G}=\left(\mathcal{V}, \mathcal{E}\right)$ to denote an attribute graph, where $\mathcal{V}= \left\{ v_1, v_2, \cdots, v_N \right\}$ is the set of $N$ nodes and $\mathcal{E} \subset \mathcal{V} \times \mathcal{V}$ is the edge set.
It can also be represented as $\mathcal{G}=\left(\mathbf{A}, \mathbf{X}\right)$. Here, $\mathbf{A} \in \left\{0, 1 \right\}^{N \times N}$ denotes the adjacency matrix where $a_{ij}=1$ if $\left( v_i, v_j \right) \in \mathcal{E}$, otherwise $a_{ij}=0$. 
$\mathbf{X} \in \mathbb{R}^{N \times d_x}$ denotes the feature matrix where the $i$-th row $\mathbf{x}_i \in \mathbb{R}^{d_x}$ is the $d_x$-dimensional feature vector of node $v_i \in \mathcal{V}$.
$\mathcal{N} \left(v_i \right)$ denotes the set of node $v_i$'s neighboring nodes.

\subsection{Graph Learning Models}
Graph learning models are powerful tools for modeling graph data by leveraging node features and graph structure to learn graph representations.
Mathematically, an $L$-layer graph learning model $f$ parameterized with $\theta$ updates the hidden representation matrix $\mathbf{H}^{\left( l \right)} \in \mathbb{R}^{N \times d_{l}}$ at the $l$-th layer by
\begin{equation} \label{gnn}
    \mathbf{H}^{\left( l \right)} = f^{\left( l \right)} \left( \mathbf{A}, \mathbf{H}^{\left( l-1 \right)}; \theta^{\left( l \right)} \right),
\end{equation}
where $\theta^{\left( l \right)}$ represents the parameters at the $l$-th layer.
The $i$-th row $\mathbf{h}^{(l)}_i$ of $\mathbf{H}^{\left( l \right)}$ denotes the $d_{l}$-dimensional hidden representation of node $v_i$ at the $l$-th layer, and we initialize $\mathbf{h}^{(0)}_i = \mathbf{x}_i$.
The output representation $\mathbf{h}^{(L)}_i$ of node $v_i$ after the final layer of the graph learning model can be used for diverse downstream tasks, such as node classification and graph classification.

\subsection{Problem Formulation}
The goal of the "pre-training, prompting" scheme is to bridge the objective gap between pre-training and downstream tasks.
To achieve this, it first trains graph learning models on pre-training tasks in an unsupervised manner, followed by adapting the pre-trained graph learning models to specific downstream tasks with trainable prompts while keeping the pre-trained graph learning models frozen.
More specifically, a graph learning model $f$ is trained through a pre-training task $\mathcal{L}_{PT}$ via self-supervised learning. 
During the prompting phase, graph prompting trains learnable prompts $\mathcal{P}$ to adapt the pre-trained graph learning model $f$ to a specific downstream task $\mathcal{L}_{DT}$.

\section{Graph Pre-training for Graph Prompting} 

Graph pre-training \citep{wang2025graph} serves as a foundational step in graph prompting by leveraging unlabeled data to capture invariance in graphs. The core objective of graph pre-training is to encode structural and semantic consistencies within graphs, thereby enhancing their utility in downstream tasks. Based on the mechanism used to derive invariance, graph pre-training methods can be broadly classified into three categories: \textit{generative}, \textit{contrastive}, and \textit{multi-task} approaches.

\subsection{Generative Methods}

Generative pre-training methods follow a denoising philosophy: portions of the input graph are deliberately corrupted, and the model is trained to recover the missing or altered information~\citep{kipf2016variational,hou2022graphmae}. Formally, the pre-training task $\mathcal{L}_{PT}$ can be defined as
\begin{equation}
  \min_{\theta} \ell(f(\tilde{\mathcal{G}}; \theta), \rho(\mathcal{G})),
\end{equation}
where $\tilde{\mathcal{G}} = \mathcal{C}(\mathcal{G})$ is the corrupted graph produced by the corruption function $\mathcal{C}(\cdot)$, and $\rho(\cdot)$ denotes the target generation function that creates reconstruction targets. The loss function $\ell(\cdot, \cdot)$ measures the discrepancy between the corrupted graph representation and the target. Based on the granularity of the generative objective, existing methods can be categorized into three levels: \textit{localized}, \textit{contextualized}, and \textit{globalized}.

\subsubsection{Localized generation}
Localized generative pre-training focuses on learning fine-grained node-level information by reconstructing missing node attributes or predicting node properties. 
For example, GraphMAE~\citep{hou2022graphmae} and following works~\citep{hou2023graphmae2,li2023s,tan2023s2gae} employ node masking, where the corruption function removes a subset of nodes $\mathcal{V}' \subset \mathcal{V}$ from the graph $\mathcal{G}$, i.e., $\mathcal{C}(\mathcal{G}) = (\mathbf{A}, \mathbf{X}_{\mathcal{V}'})$, and the target generation function aims to reconstruct them $\rho(\mathcal{G}) = \mathbf{X}_{\mathcal{V}'}$. 
Alternatively, other methods~\citep{hu2019strategies} mask node feature attributes i.e., $\mathcal{C}(\mathcal{G}) = (\mathbf{A}, \tilde{\mathbf{X}})$, and train the model to predict their original values $\rho(\mathcal{G}) = \mathbf{X}$. 
Additionally, several methods aim to predict intrinsic node properties, such as centralities~\citep{hu2019pre} and clustering coefficients~\citep{jin2020self}. 
In these cases, corruption may be an identity function $\mathcal{C}(\mathcal{G}) = (\mathbf{A}, \mathbf{X})$ (i.e., no corruption), while $\rho(\mathcal{G}) = p$ generates the desired statistical measures $p$.

\subsubsection{Contextualized generation}
Contextualized generative methods predict structural information surrounding a node, thereby capturing local neighborhood dependencies. Graph autoencoders (GAE)~\citep{kipf2016variational} reconstruct graph topology by corrupting edges and predicting their original connectivity. Here, $\mathcal{C}(\mathcal{G}) = (\tilde{\mathbf{A}}, \mathbf{X})$ introduces edge removing or perturbations, and $\rho(\mathcal{G}) = \mathbf{A}$ recovers the missing or altered edges. The link prediction task also serves as a fundamental component in certain graph prompting designs~\citep{sun2022gppt,liu2023graphprompt,yu2024generalized}. Beyond edge prediction, contextualized generation has been extended to capture higher-order structures. Some methods predict shortest paths~\citep{peng2020self} between node pairs to preserve long-range dependencies, while others focus on identifying subgraph motifs~\citep{rong2020self} or semantic clusters~\citep{sun2020multi,you2020does} relevant to specific target nodes.

\subsubsection{Globalized generation}
Globalized generative methods pre-train graph learning models by preserving entire graph-level properties. In these methods, the graph often remains unchanged; therefore, the corruption function $\mathcal{C}(\mathcal{G}) = (\mathbf{A}, \mathbf{X})$ is the identity function. 
A graph learning model can be pre-trained through graph similarity-based objectives, where the representations of two graphs generated by the graph learning model are compared using metrics such as graph kernel similarities \citep{navarin2018pre}, graph edit distances \citep{kim2022graph}, isomorphism scores \citep{li2024hierarchical}, or hyperbolic distances \citep{skenderi2023graph}.

\subsection{Contrastive Methods}
Inspired by the advances of contrastive learning in computer vision and NLP, recent studies apply contrastive learning to graph data for self-supervised graph pre-training. 
Basically, contrastive methods pre-train graph learning models by differentiating similar and dissimilar instances in graph data. 
First, multiple transformations are applied to generate different augmented views.
More specifically, given a graph $\mathcal{G}=\left(\mathbf{A}, \mathbf{X}\right)$, $K$ transformations $\mathcal{T}_1, \mathcal{T}_2, \dots, \mathcal{T}_K$ generate $K$ different augmented views by
\begin{equation}
    \tilde{\mathcal{G}_k}=\left(\mathbf{A}_k, \mathbf{X}_k\right)=\mathcal{T}_K\left(\mathbf{A}, \mathbf{X}\right), \; k = 1, 2, \cdots, K
\end{equation}
$K$ graph learning models (independent or weight-sharing) then map these views into representations $\left\{\mathbf{H}^{\left( L \right)}_1,\mathbf{H}^{\left( L \right)}_2,\cdots,\mathbf{H}^{\left( L \right)}_K \right\}$, where each representation $\mathbf{H}^{\left( L \right)}_k = f \left(\tilde{\mathcal{G}_k}; \theta_k \right)$. 
The objective of contrastive learning is to maximize the similarity between positive pairs while minimizing the agreement between negative pairs by
\begin{equation}
    \begin{aligned}
    \max_{ \left\{\theta_k\right\}_{k=1}^K} & \sum_{i\neq j}s_{ij}\mathcal{I}(\mathbf{h}_i, \mathbf{h}_j), \\
     \text{where} \; \mathcal{I}(\mathbf{h}_i, \mathbf{h}_j) = & \; \mathbb{E}_{p(\mathbf{h}_i, \mathbf{h}_j)}\left[\log\frac{p(\mathbf{h}_i, \mathbf{h}_j)}{p(\mathbf{h}_i)p(\mathbf{h}_j)}\right].
\end{aligned}
\end{equation}
Here, $\mathcal{I}(\cdot, \cdot)$ represents the mutual information (MI), and $s_{ij} \in \{0,1\}$ indicates whether $\mathcal{I}$ is computed between $\mathbf{h}_i$ and $\mathbf{h}_j$.
$p(\mathbf{h}_i, \mathbf{h}_j)$ denotes the joint distribution of $\mathbf{h}_i$ and $\mathbf{h}_j$, and $p(\mathbf{h}_i)p(\mathbf{h}_j)$ denotes the product of marginals. 
Since the MI is intractable, contrastive methods usually pre-train graph learning models by minimizing contrastive loss, such as InfoNCE~\citep{gutmann2010noise} and Jensen-Shannon divergence~\citep{nowozin2016f}.
Generally, contrastive methods incorporate two key components for graph pre-training: \emph{graph augmentation}, and \emph{contrastive pair selection}.

\definecolor{myred}{RGB}{215,48,39}
\definecolor{mygreen}{RGB}{26,152,80}
\newcommand{\cmark}{\textcolor{mygreen}{\ding{51}}}
\newcommand{\xmark}{\textcolor{myred}{\ding{55}}}

\begin{table*}[t]
\small
\centering
\caption{A summary of graph prompting techniques in the existing studies. \textit{Single Forward Pass} means graph learning models only perform the forward once during graph prompting. \textit{DT Universality} means graph prompting techniques are universal for different downstream tasks, including classification and regression.
}
\label{table:table_methods}
\setlength\tabcolsep{3.01pt}
\setlength{\extrarowheight}{2pt}

\setlength\dashlinegap{1.5pt}

\begin{tabular}{cccccccccc}
\rowcolor{lightgray!30}\specialrule{1pt}{0pt}{0pt}
\bf{Technique Categories}   & \bf{Prompting Stage}                  & \bf{Prompting Strategies}                                     & \bf{Single Forward Pass}                                  & \bf{DT Universality}         & \bf{Related Works}                 \\ \specialrule{1pt}{0pt}{0pt}
\multirow{6}{*}{\makecell{\bf{Data-level}\\ \bf{prompting}}}        &  \multirow{6}{*}{Graph data}                                  & \multirow{4}{*}{\makecell{Feature-based\\ prompting}}     & \multirow{4}{*}{\xmark}           & \multirow{4}{*}{\cmark}       & GPF-plus~\cite{fang2023gpf}, IAGPL~\cite{li2025instanceaware},    \\
                            &                                       &                                                               &                                           &                               & GSPF~\cite{jiang2024unified}, HetGPT~\cite{ma2024hetgpt},             \\
                            &                                       &                                                               &                                           &                               & SUPT~\cite{lee2024subgraph}, DyGPrompt~\cite{yu2025nodetime},         \\
                            &                                       &                                                               &                                           &                               & FPrompt~\cite{li2025fairnessaware}, RELIEF~\cite{zhu2024relief}       \\ \cdashline{3-6}
                            &                                       & \multirow{2}{*}{\makecell{Insertion-based \\ prompting}}      & \multirow{2}{*}{\xmark}                   & \multirow{2}{*}{\cmark}       & PSP~\citep{ge2024psp}, All-in-one~\cite{sun2023all},                  \\
                            &                                       &                                                               &                                           &                               & VNT~\cite{tan2023vnt}, SGL-PT~\cite{zhu2023sgl}                       \\ 
                            \specialrule{0.7pt}{0pt}{0pt}
                            &                                       & \multirow{2}{*}{\makecell{Output representation\\ prompting}} & \multirow{2}{*}{\cmark}                   & \multirow{2}{*}{\cmark}       & GraphPrompt~\cite{liu2023graphprompt},                                \\
\bf{Representation-level}   & \multirow{2}{*}{Node representations} &                                                               &                                           &                               & ProNoG~\cite{yu2025pronog}                                            \\                                     \cdashline{3-6}
\bf{prompting}              &                                       & \multirow{2}{*}{\makecell{Hidden representation\\ prompting}} & \multirow{2}{*}{\xmark}                   & \multirow{2}{*}{\cmark}       & EdgePrompt+~\cite{fu2025edge},                                        \\
                            &                                       &                                                               &                                           &                               & GraphPrompt+~\cite{yu2024generalized}                                 \\ 
                            \specialrule{0.7pt}{0pt}{0pt}
\multirow{5}{*}{\makecell{\bf{Task-level}\\ \bf{prompting}}}        & \multirow{5}{*}{Downstream tasks}     & \multirow{2}{*}{\makecell{Link prediction-based \\ prompting}}    & \multirow{2}{*}{\cmark}       & \multirow{2}{*}{\xmark}       & \multirow{2}{*}{GPPT~\cite{sun2022gppt}} \\
                            &                                       &                                                               &                                           &                               &     \\ 
                            \cdashline{3-6}
                            &                                       & \multirow{3}{*}{\makecell{Similarity-based \\ prompting}}     & \multirow{3}{*}{\cmark}                   & \multirow{3}{*}{\xmark}       & ULTRA-DP~\cite{chen2023ultra}, GraphPrompt~\cite{liu2023graphprompt}, \\
                            &                                       &                                                               &                                           &                               & GraphPrompt+~\cite{yu2024generalized},  HetGPT~\cite{ma2024hetgpt},   \\
                            &                                       &                                                               &                                           &                               & MultiGPrompt~\citep{yu2024multigprompt}, ProNoG~\cite{yu2025pronog}   \\
\specialrule{1pt}{0pt}{0pt}
\end{tabular}
\end{table*} 

\subsubsection{Graph augmentation}
The goal of graph augmentation is to generate augmented graphs that can enrich or enhance the preserved information from the given graph data.
One prevalent technique in graph augmentation is \emph{feature-based augmentation}. 
For instance, node feature masking~\citep{jin2020self,hu2019strategies,you2020graph,zhu2020deep,thakoor2021large,jin2021multi} randomly select a portion of node attributes and replace them with constant or some predefined distribution.
Similarly, node feature shuffling~\citep{velickovic2019deep,opolka2019spatio,ma2021calibrating,jing2021hdmi,ren2019heterogeneous} switch features among a small fraction of nodes. 
In the meantime, \emph{structure-based augmentation} has been adopted by many studies via transformation on graph topology.
For example, edge perturbation~\citep{zhu2020self,you2020graph,hu2020gpt,zhang2023iterative,zeng2021contrastive,jin2021multi,qiu2020gcc} randomly adds or removes a portion of edges in $\mathcal{G}$, while edge diffusion~\citep{hassani2020contrastive, kefato2021self, gasteiger2019diffusion} augments $\mathbf{A}$ by incorporating global topological information, linking nodes to their indirectly connected neighbors with computed weights. 
In addition, some studies also use \emph{sampling-based augmentation} that samples subgraphs of $\mathcal{G}$ with a node set $\mathcal{V}'\subseteq \mathcal{V}$. Common sampling strategies include uniform sampling~\citep{zeng2021contrastive}, ego-net sampling~\citep{hu2019strategies,zhu2021transfer,cao2021bipartite}, random walk sampling~\citep{hassani2020contrastive,qiu2020gcc,you2020graph}, and others~\citep{jiao2020sub,zhang2024motif,wang2021self}.

\subsubsection{Contrastive pair selection}
The second component of contrastive methods is to select appropriate positive and negative pairs.
Typically, contrastive pairs are either the same-scale or cross-scale.
The same-scale pairs are selected at the same granularity.
For instance, (sub)graph-level methods~\citep{you2020graph, zeng2021contrastive, you2021graph, suresh2021adversarial, verma2021towards, qiu2020gcc} pre-train graph learning models to predict whether two (sub)graphs originate from the same source, while node-level methods~\citep{hwang2020self, tang2015line,hamilton2017inductive, grover2016node2vec, perozzi2014deepwalk} align node representations more closely with their contextual neighborhoods.
Modern approaches~\citep{zhu2020deep, thakoor2021large, kefato2021self, opolka2019spatio, peng2020graph, yu2021socially, jovanovic2021towards, zhu2021graph, wang2021self, wan2021contrastive} extend this idea by incorporating views from augmented graphs. 
In contrast, cross-scale pairs focus on modeling hierarchical relationships in graph data.
For instance, node-(sub)graph methods~\citep{ren2019heterogeneous, li2021leveraging, park2020unsupervised, opolka2019spatio, zhu2021transfer, wang2021self, velickovic2019deep, hassani2020contrastive, jiao2020sub, mavromatis2020graph} contrast node representations with (sub)graph representations, aiming to maximize the MI between them. This encourages graph learning models to capture both localized and global semantics. 
Subgraph-global methods~\citep{zhang2024motif, cao2021bipartite, sun2019infograph, wang2021learning, sun2021sugar}, on the other hand, sample subgraphs from the original or augmented graphs and optimize MI between local contextual and global representations.

\subsection{Multi-task Methods}
Beyond prevalent generative and contrastive methods, several studies propose to pre-train graph learning models through multi-task learning.
For instance, CPT-HG~\citep{jiang2021contrastive} designs a contrastive pre-training strategy for GNNs on heterogeneous graphs, which leverages both relation-level and subgraph-level pre-training tasks.
Hu et al.~\citeN{hu2019strategies} investigate pre-training GNNs using node-level tasks (e.g., context prediction and attribute masking) and graph-level tasks (e.g., graph property prediction and structural similarity prediction).
MultiGPrompt~\citep{yu2024multigprompt} introduces a set of pretext tokens for multiple pre-training tasks during pre-training, which are later used as composed prompts for adaptation.

\section{Techniques in Graph Prompting} 
In this section, we introduce existing techniques in graph prompting.
Based on different stages in downstream tasks, these techniques can be roughly divided into three categories, namely data-level prompting, representation-level prompting, and task-level prompting.
Table~\ref{table:table_methods} summarizes graph prompting techniques in different categories.
For the techniques in each category, we will elaborate on their detailed design of graph prompts for adapting pre-trained graph learning models.

\subsection{Data-level Prompting}
Inspired by prevalent prompting methods in natural language processing~\cite{zhou2022coop,khattak2023maple}, a straightforward graph prompting strategy is data-level prompting. 
The intuition behind data-level prompting is to modify the input graph data to fit specific downstream tasks.
Specifically, given the input graph $\mathcal{G} = (\mathbf{A}, \mathbf{X})$, data-level prompting $\mathcal{T}^D$ learns to transform it to a prompted graph $\tilde{\mathcal{G}} = (\tilde{\mathbf{A}}, \tilde{\mathbf{X}})$ with prompts $\mathcal{P}$.
Formally, we can formulate the above technique by
\begin{equation}
    (\tilde{\mathbf{A}}, \tilde{\mathbf{X}} ) = \mathcal{T}^D \left(\left(\mathbf{A}, \mathbf{X}\right); \mathcal{P}\right).
\end{equation}
It is worth noting that $\mathcal{T}^D$ may not apply $\mathcal{P}$ to both $\mathbf{A}$ and $\mathbf{X}$ simultaneously.
According to the modified components, data-level prompting can be categorized into two types: feature-based prompting and insertion-based prompting.

\subsubsection{Feature-based prompting}
We first introduce feature-based prompting by solely modifying the feature matrix.
In this case, the graph topology remains unchanged, i.e., $\tilde{\mathbf{A}} = \mathbf{A}$.
Basically, feature-based prompting learns an additional prompt vector that is added to the feature vector of each node~\cite{fang2023gpf, lee2024subgraph, li2025instanceaware, zhu2024relief, jiang2024unified, li2025fairnessaware, ma2024hetgpt, yu2025nodetime}.
Given a learnable prompt vector $\mathbf{p}_i \in \mathbb{R}^{d_x}$, node $v_i$ will have a prompted feature vector $\tilde{\mathbf{x}}_i = \mathbf{x}_i + \mathbf{p}_i$. 
The major challenge here is how to formulate the prompt vector $\mathbf{p}_i$.
GPF~\cite{fang2023gpf} overcomes the challenge by simply learning a prompt vector $\mathbf{p} \in \mathbb{R}^{d_x}$ shared by all the nodes, i.e., $\mathbf{p}_1=\mathbf{p}_2=\cdots=\mathbf{p}_N=\mathbf{p}$.
Its variant GPF-plus~\cite{fang2023gpf} further assigns an independent prompt vector to each node. 
To handle large-scale graphs, the attentive aggregation of multiple basis vectors is employed to generate the prompt vector, making GPF-plus more parameter-efficient and capable of handling graphs with different scales.
More specifically, given a set of $M$ $d_x$-dimensional basis vectors $\mathbf{b}_1, \mathbf{b}_2, \cdots, \mathbf{b}_M$, we compute the prompted feature vector for each node $v_i$ by
\begin{equation}
    \tilde{\mathbf{x}}_i = \mathbf{x}_i + \mathbf{p}_i = \mathbf{x}_i + \sum_{m=1}^M w_{i,m} \cdot \mathbf{b}_m,
\end{equation}
where each $w_{i,m}$ is the weight value in the weight vector $\mathbf{w}_i \in \mathbb{R}^{M}$ of node $v_i$. In GPF-plus, $\mathbf{w}_i$ is computed as the softmax values of projected node feature $\mathbf{x}_i$.
Its following work SUPT~\citep{lee2024subgraph} computes $\mathbf{w}_i$ using the resulting scores derived from simple GNNs, while IAGPL~\cite{li2025instanceaware} introduces node-level instance-aware graph prompt vectors to obtain $\mathbf{w}_i$.
It utilized a parameter-efficient backbone model to transform the feature space into the prompt space so that these generated prompts are node-aware. 
RELIEF~\cite{zhu2024relief} points out that graph prompts are necessary for specific nodes.
Instead of learning prompt vectors for each node, RELIEF formulates the process of inserting prompts as a sequential decision-making problem and designs a reinforcement learning framework to select which node to prompt.
GSPF~\cite{jiang2024unified} proposes to assign an additional learnable prompt weight to adjust the influence of prompt vectors to each node.
FPrompt~\cite{li2025fairnessaware} focuses on the issue of fairness in feature-based prompting.
To promote fairness through graph prompting, it proposes to incorporate sensitivity feature(s) into prompt vectors.

Feature-based prompting can also be adapted for other types of graphs.
For instance, HetGPT~\cite{ma2024hetgpt} generalizes feature-based prompting on heterogeneous graphs.
Considering multiple node types in heterogeneous graphs, it designs type-wise weight vectors with specific basis vectors for each node type.
DyGPrompt~\cite{yu2025nodetime} focuses on feature-based prompting for dynamic graphs. 
It investigates the gaps arising from both task and dynamic differences between pre-training and downstream tasks.
To bridge the gap, it employs dual conditional networks~\cite{zhou2022conditional} to generate dual prompts from the perspectives of nodes and time, respectively.

\subsubsection{Insertion-based prompting}
The intuition of insertion-based prompting is to insert additional prompt nodes as learnable prompts into the original graph~\cite{sun2023all, tan2023vnt, zhu2023sgl, ge2024psp}. As a result, the original graph $\mathcal{G}$ is transformed into a prompted graph $\tilde{\mathcal{G}}$ where the nodes include the prompt nodes and those of the original graph.
In this case, the learnable prompts in $\mathcal{P}$ are $M$ prompt vectors, i.e.,  $\mathcal{P} = \left\{ \mathbf{p}_1, \mathbf{p}_2, \cdots, \mathbf{p}_M \right\}$.
Compared to feature-based prompting, the primary challenge here lies in appropriately integrating the prompt nodes into the original graph. 
SGL-PT~\cite{zhu2023sgl} follows the concept of virtual nodes~\cite{hu2020ogb, hu2021ogb} in graph learning by uniformly connecting one prompt node with every node in the original graph.
Similarly, PSP~\citep{ge2024psp} designs multiple prompt nodes that are uniformly connected to all graph nodes.
These prompt nodes are treated as the class prototypes for classification tasks.
All-in-one~\cite{sun2023all} defines the inserting pattern using the dot product between prompt nodes and the original graph nodes.
VNT~\cite{tan2023vnt} particularly focuses on the inserting pattern for pre-trained graph transformers. Like prompt designs in language models~\cite{zhou2022learning, zhou2022conditional}, it concatenates the prompt nodes with the graph nodes and feeds them to the transformer layers of the pre-trained graph transformers.
Another challenge in insertion-based prompting is determining how to appropriately connect the prompt nodes to each other, especially when inserting multiple prompt nodes.
All-in-one~\cite{sun2023all} solves this challenge by proposing three methods: 1) learn free parameters indicating how possible the prompt nodes should be connected; 2) compute the dot product of each pair of prompt nodes and prune it with a pre-defined threshold; 3) treat the prompt nodes as independent and unconnected.

While data-level prompting is straightforward and effective for various downstream tasks, it requires performing the entire forward and backward pass of the pre-trained graph learning models for each update.
Compared to other graph prompting techniques (e.g., task-level prompting), it can cause significant computational costs to train graph prompts and time delay during inference.

\subsection{Representation-level Prompting}
Beyond learning to modify the input graph data, another key graph prompting technique is representation-level prompting, which designs and applies learnable prompts to node representations.
Here, node representations can refer to the output representations after the final layer or the hidden representations at each layer of pre-trained graph learning models. 
Specifically, given the (hidden) representations matrix $\mathbf{H}^{\left( l \right)}$ at the $l$-th layer, representation-level prompting $\mathcal{T}^R$ learns to transform it to a prompted representations matrix $\tilde{\mathbf{H}}^{\left( l \right)}$ with prompts $\mathcal{P}$.
Mathematically, we can formulate representation-level prompting $\mathcal{T}^R$ by
\begin{equation}
    \tilde{\mathbf{H}}^{\left( l \right)} = \mathcal{T}^R \left(\mathbf{H}^{\left( l \right)}; \mathcal{P}\right).
\end{equation}
When the prompts are applied only to the output representations, we have $l = L$. 
According to the layers of pre-trained graph learning models to which prompt vectors are applied, representation-level prompting methods can be categorized into \emph{output representation prompting} and \emph{hidden representation prompting}.

\subsubsection{Output representation prompting}
In output representation prompting, prompt vectors directly modify the output representations after the final layer of pre-trained graph learning models~\cite{liu2023graphprompt, yu2025pronog}.
The modified representations are then used for downstream tasks.
Typically, output representation prompting methods employ a learnable prompt vector $\mathbf{p}_i \in \mathbb{R}^{d_{l}}$ for each node $v_i$. 
Mathematically, given the output representation $\mathbf{h}^{\left( L \right)}_i$ of node $v_i$, its modified representation $\tilde{\mathbf{h}}^{\left( L \right)}_i$ is obtained through an element-wise multiplication by
\begin{equation}
    \tilde{\mathbf{h}}^{\left( L \right)}_i = \mathbf{p}_i \odot \mathbf{h}^{\left( L \right)}_i,
\end{equation}
where $\odot$ denotes the element-wise multiplication.
The prompt vector $\mathbf{p}_i$ here serves as dimension-wise weights that extract the most relevant prior knowledge for downstream tasks.
Intuitively, GraphPrompt~\citep{liu2023graphprompt} uses a universal prompt vector shared by all the nodes, i.e., $\mathbf{p}_1=\mathbf{p}_2=\cdots=\mathbf{p}_N$.
To perform fine-grained adaptation, ProNoG~\citep{yu2025pronog} trains a conditional network~\citep{zhou2022conditional} to generate node-wise prompt vectors based on the output representations of nodes within the multi-hop ego-network of each target node.
This strategy is particularly important for heterophilic graphs because the non-homophilic patterns of a node are mainly characterized by its multi-hop neighborhood.

\subsubsection{Hidden representation prompting}
Unlike output representation prompting where prompt vectors are applied to output representations, hidden representation prompting methods design layer-wise prompts for pre-trained graph learning models~\cite{yu2024generalized, fu2025edge}.
For instance, GraphPrompt+~\cite{yu2024generalized} enhances GraphPrompt by modifying hidden representations at each layer of pre-trained graph learning models. 
Therefore, it will learn $L+1$ prompt vectors, with one prompt vector allocated for each layer, including the input layer.
Specifically, the modified hidden representation is computed by $\tilde{\mathbf{h}}^{\left( l \right)}_i = \mathbf{p}^{\left( l \right)}_i \odot \mathbf{h}^{\left( l \right)}_i$.
Additionally, EdgePrompt~\citep{fu2025edge} proposes to learn a shared prompt vector for each edge at each layer of pre-trained graph learning models.
Its variant EdgePrompt+~\citep{fu2025edge} extends the idea of edge prompts by learning customized prompt vectors for each edge.
Specifically, given each edge $\left( v_i, v_j \right) \in \mathcal{E}$, its customized prompt vector $\mathbf{e}^{\left( l \right)}_{i,j}$ at the $l$-th layer is computed based on the hidden representations $\mathbf{h}^{\left( l \right)}_i$ and $\mathbf{h}^{\left( l \right)}_j$ of nodes $v_i$ and $v_j$.
The edge prompt $\mathbf{e}^{\left( l \right)}_{i,j}$ is then aggregated along with hidden representations of nodes through the message-passing mechanism during the forward pass at each layer of pre-trained graph learning models.

Compared to output representation prompting, hidden representation prompting further enhances adaptation through layer-wise prompt vectors.
However, it requires hidden representations at each layer of pre-trained graph learning models, which may be inapplicable to black-box graph learning models.
In contrast, output representation prompting only needs the ultimate outputs of pre-trained graph learning models, making it more flexible and computationally efficient.

\subsection{Task-level Prompting}
During the adaptation stage, graph prompting often encounters the few-shot setting, i.e., pre-trained graph learning models are adapted using a limited amount of labeled graph data.
A straightforward strategy to address this scenario is to train linear probes~\cite{fang2023gpf, lee2024subgraph, fu2025edge} using traditional loss functions, such as cross-entropy loss for classification tasks and mean square error for regression tasks.
Beyond training linear probes, numerous studies explore reformulating downstream tasks---particularly classification tasks---into alternative forms.
We term this technique task-level prompting.
More specifically, given a downstream task $\mathcal{L}_{DT}$ (e.g., node classification), task-level prompting $\mathcal{T}^T$ transforms it into a different task $\tilde{\mathcal{L}}_{DT}$ for adaptation.
It is worth noting that $\mathcal{T}^T$ may incorporate additional learnable prompts $\mathcal{P}$ in some cases.
Mathematically, we can formulate task-level prompting by
\begin{equation}
    \tilde{\mathcal{L}}_{DT} = \mathcal{T}^T \left(\mathcal{L}_{DT}; \mathcal{P}\right).
\end{equation}
Generally, task-level prompting methods can be roughly categorized into two types: \emph{link prediction-based prompting} and \emph{similarity-based prompting}.

\subsubsection{Link prediction-based prompting}
The objective gap arises when link prediction is used to pre-train graph learning models, which are then applied to node classification as the downstream task.
Link prediction-based prompting~\citep{sun2022gppt} bridges this gap by converting node classification into link prediction. Therefore, we have $\tilde{\mathcal{L}}_{DT} = \mathcal{L}_{PT}$.
More specifically, link prediction-based prompting enables each class $c \in \mathcal{Y}$ to have a learnable prompt vector $\mathbf{s}_c$. $\mathbf{s}_c$ is combined with the representation $\mathbf{h}_i$ of a target node $v_i$ as the input of the projection head pre-trained by link prediction. 
If node $v_i$'s ground-truth label $y_i=c$, the projection head will treat node $v_i$ and class $c$ as "a pair of connected nodes".
Therefore, the prompt vectors are tuned to predict the highest linking probability between $\mathbf{h}_i$ and $\mathbf{s}_c$ by the projection head.
Considering the cluster structure of graph data, GPPT~\citep{sun2022gppt} proposes to learn cluster-wise prompt vectors for each class.
Although link prediction-based prompting is very straightforward, it is only compatible with link prediction as the pre-training task since it requires the projection head pre-trained by link prediction.
In addition, it only handles node classification as the downstream task and is inapplicable to other tasks, such as node regression and graph classification.

\subsubsection{Similarity-based prompting}
Instead of using cross-entropy loss for classification tasks, similarity-based prompting compares graph representations to class prototypes using contrastive loss~\citep{liu2023graphprompt, yu2024generalized, yu2024multigprompt, yu2024non, ma2024hetgpt}.
More specifically, given an instance-label pair $(x, y)$ in the training set $\mathcal{D}$, where $x$ is an instance that can be either a node or a graph, and $y \in \mathcal{Y}$ is its ground-truth class label from a set of classes $\mathcal{Y}$, similarity-based prompting aims to maximize the similarity between $x$'s representation $\mathbf{h}_x$ and the class prototype $\mathbf{s}_y$ of class $y$.
Formally, we formulate the similarity task $\tilde{\mathcal{L}}_{DT}$ in similarity-based prompting by
\begin{equation}
    \min_{\mathcal{P}} \sum_{\left( x, y \right) \in \mathcal{D}} - \log \frac{\exp\left(\text{sim} \left( \mathbf{h}_x, \mathbf{s}_y \right) / \tau \right)}{\sum_{c \in \mathcal{Y}} \exp\left(\text{sim} \left( \mathbf{h}_x, \mathbf{s}_c \right) / \tau \right)},
\end{equation}
where $\text{sim} \left( \cdot, \cdot \right)$ denotes a similarity function between two vectors, and $\tau$ denotes a temperature hyperparameter to control the shape of the output distribution.
To obtain the class prototype $\mathbf{s}_c$, a few studies use the average of instance representations belonging to class $c$~\citep{liu2023graphprompt,yu2024generalized,yu2024multigprompt, yu2024non}.
Therefore, no additional prompts are introduced in similarity-based prompting.
In the meantime, a few studies design class prototypes as learnable vectors~\citep{ma2024hetgpt, chen2023ultra}.
In this case, class prototypes are also included in the learnable prompts.

While task-level prompting methods can effectively adapt pre-trained graph learning models by converting downstream tasks into other tasks (e.g., from link prediction to node classification), their key limitation is that they can only handle classification tasks. In other words, current task-level prompting methods are ineffective when applied to tasks other than classification, such as regression.

\section{Applications} 
Graph prompting has been widely applied in various domains, including recommendation systems, knowledge engineering, biology, and medicine.
In this section, we introduce some representative real-world applications of graph prompting.
\subsection{Recommendation Systems}
Prompting techniques have been widely adopted in language model-based recommendation systems to enhance personalization and explanations~\cite{geng2022recommendation, liu2023pre, zhu2024collaborative}. 
Since user-item interactions naturally form a bipartite graph, GNN-based recommenders such as LightGCN~\cite{he2020lightgcn} have also gained significant attention, and graph prompting can be introduced to further enhance their adaptivity. 
For example, PGPRec~\cite{yi2023contrastive} introduces domain-specific node prompts to adapt user/item embeddings in GNN-based recommenders to another domain where items could be different. 
GraphPro~\cite{yang2024graphpro} captures users' evolving interests by pre-training GNN-based recommenders with the relative time of interactions and using edges from a specific time as edge prompts to dynamically adjust user preferences. 
CPTPP~\cite{yang2023empirical} uses the interaction matrix to infer personalized prompts that can be augmented with pre-trained user embeddings for recommendations. 
GPT4Rec~\cite{zhang2024gpt4rec} models the temporal evolution of users and items by introducing multiple hidden views of the interaction graph and leveraging both node and edge prompts to capture these changes. 
More recently, graph prompting has been generalized to graph language model-based recommendation systems, where graphs are transformed into token sequences and enriched with textual prompts tailored for different downstream tasks~\cite{zhu2024understanding, wei2024llmrec}.

\subsection{Knowledge Engineering}
Another important application is knowledge engineering, as real-world knowledge forms knowledge graphs, and graph prompting facilitates efficient information retrieval with good generalization ability across different knowledge graphs and downstream tasks. 
KGTransformer~\citep{zhang2023structure} pre-trains graph learning models using self-supervised learning on sampled subgraphs, where the task triplet serves as a prompt to enable flexible interactions between task-specific data and KG representations. 
MUDOK~\citep{zhang2024multi} pre-trains KG models through multi-domain collaborative pre-training and efficient prefix prompting, where the task triplet serves as a prompt to enhance transferability across applications. 
KG-ICL~\citep{cui2025prompt} utilizes a prompt-based approach to construct query-specific prompt graphs, enabling the model to generalize across different knowledge graphs by leveraging in-context learning to perform universal relational reasoning without fine-tuning.

\subsection{Biology and Medicine}
Graph prompting also found applications in biology and medicine, where graph-structured data are prevalent while labels are very scarce. 
One key application is molecular graph modeling, where atoms are represented as nodes and bonds as edges. 
Given pre-trained molecular graph models, MolCPT~\cite{diao2022molcpt} infers continuous node prompts from molecular motifs that are predictive of molecular properties. 
MOAT~\cite{long2024moat}employs geometry-level prompts that can be adapted for molecular conformers. 
TGPT~\cite{wang2024novel}leverages graphlets and their frequencies to infer node-, graph-, and task-level graph prompt embeddings, which improve the adaptability of graph molecule models to various molecular topologies. 
Beyond molecular graphs, graph prompting has also been applied to drug interaction event prediction~\cite{wang2024ddiprompt}, where molecular structures and their interconnections are used for pre-training graph learning models, while trainable graph prompts that indicate interaction prototypes are introduced to adapt the model to different event types. 
Furthermore, in medical imaging applications, MMGPL~\cite{peng2024mmgpl} utilizes MRI and PET data to learn node prompts that can be combined with pre-trained disease concept graphs to enhance diagnostic accuracy.

\section{Future Directions} 
In this section, we present some limitations in current studies and provide promising directions for future advances.


\subsection{Benchmark Construction}
Despite the effectiveness of graph prompting techniques, existing evaluation datasets face several critical limitations. 
First, current graph prompting studies rely on disparate datasets for evaluation, which makes it difficult to ensure fair comparisons.
Second, empirical evaluations are conducted in different settings, including variations in the number of shots, batch sizes, learning rates, and other factors.
Third, many graph prompting methods are not evaluated with sufficient pre-training strategies, which makes their compatibility unclear.
So far, only one benchmark~\citep{zi2024prog} has been introduced to standardize the evaluation of graph prompting methods.
However, the performance of graph prompting techniques in more settings remains underexplored.

\subsection{Theoretical Foundation}
While graph prompting has achieved empirical success by adapting concepts from large language models (LLMs)~\citep{gong2023prompt, fang2023gpf}, its theoretical underpinnings remain critically underdeveloped. Current methods lack rigorous analysis of why prompts enable effective knowledge transfer or how their structure aligns with intrinsic graph properties, which not only limits explainability but also creates performance bottlenecks. Several works have made early explorations for theoretical groundings~\citep{wang2024does,huang2024graph, fang2023universal}. Future work can further bridge this gap by establishing formal theoretical frameworks grounded in graph theory (e.g. spectral graph theory and geometric deep learning) to systematically guide prompt design, unlocking interpretable and robust performance.

\subsection{Universal Compatibility}
Unlike NLP prompts which benefit from the structural consistency of language to guide LLMs across diverse tasks, existing graph prompts are tailored to the specific characteristics of individual graphs and tasks~\citep{liu2023graphprompt, sun2022gppt}. In addition, there are few explorations of prompt designs for graph transformers which have better expressive ability~\citep{tan2023vnt}. As a result, it would be helpful to develop unified graph prompting frameworks capable of adapting to diverse graph structures and tasks and compatible with graph transformers without manual tailoring. Possible directions include utilizing dynamic prompt architectures that can self-align with graph properties (e.g., scale and sparsity). Additionally, integrating cross-domain alignment mechanisms~\citep{chen2020graph, zhao2023cross} could bridge gaps between disparate graph spaces, facilitating the universal applicability of graph prompting.


\subsection{Model Robustness}
The robustness of graph prompting against adversarial attacks remains understudied, with existing methods lacking strategies and rigorous analysis against adversarial perturbations, which presents a promising research direction. First, graph prompting can enhance GNN robustness by addressing structural perturbations~\citep{zugner2020certifiable, wang2021certified} (e.g., node injections and edge modifications) and feature-based attacks~\citep{chen2023feature, xu2024attacks} (e.g., adversarial noise) on GNNs. Second, graph prompting itself requires defense, as current node-, edge-, or feature-centric prompt designs risk amplifying structural or feature vulnerabilities in the underlying model~\citep{lyu2024cross, lin2024trojan}. Future work can prioritize rigorous adversarial testing and integrate defense mechanisms to ensure the reliability of graph prompting in adversarial settings.



\subsection{LLM Incorporation}
While graph foundation models are limited by parameter size~\citep{sun2023survey}, limiting their ability to fully exploit graph prompting, LLMs offer complementary strengths through exceptional in-context reasoning ability~\citep{brown2020language, lester2021power}. Current approaches bridge this gap by either grounding graph prompts into language space for LLM processing~\citep{ye2023language, li2024graph, tian2024graph} or using LLMs to generate semantic label prompts that enhance task alignment~\citep{duan2024g}. However, broader opportunities remain unexplored, such as leveraging LLMs to design topology-aware prompts and prompt optimization through iterative feedback with LLMs, which can further enhance graph prompting with the powerful in-context learning and reasoning abilities of LLMs.





\section{Conclusion} 
In this survey, we provide a comprehensive review of recent techniques in graph prompting for graph learning models. 
We first introduce representative graph pre-training methods that serve as the foundation stage of graph prompting.
Then, we dive deep into graph prompting techniques from the perspectives of graph data, node representation, and downstream tasks.
Moreover, we present some applications of graph prompting in various domains, including recommendation systems, knowledge engineering, biology, and medicine.
Finally, we discuss critical limitations in the existing graph prompting studies with their corresponding possible directions in the future.

\begin{acks}
This work is supported in part by the National Science Foundation (NSF) under grants IIS-2006844, IIS-2144209, IIS-2223769, CNS-2154962, BCS-2228534, and CMMI-2411248; the Office of Naval Research (ONR) under grant N000142412636; and the Commonwealth Cyber Initiative (CCI) under grant VV-1Q24-011.
\end{acks}

\bibliographystyle{ACM-Reference-Format}
\balance 
\bibliography{reference}


\begin{thebibliography}{151}


\ifx \showCODEN    \undefined \def \showCODEN     #1{\unskip}     \fi
\ifx \showISBNx    \undefined \def \showISBNx     #1{\unskip}     \fi
\ifx \showISBNxiii \undefined \def \showISBNxiii  #1{\unskip}     \fi
\ifx \showISSN     \undefined \def \showISSN      #1{\unskip}     \fi
\ifx \showLCCN     \undefined \def \showLCCN      #1{\unskip}     \fi
\ifx \shownote     \undefined \def \shownote      #1{#1}          \fi
\ifx \showarticletitle \undefined \def \showarticletitle #1{#1}   \fi
\ifx \showURL      \undefined \def \showURL       {\relax}        \fi
\providecommand\bibfield[2]{#2}
\providecommand\bibinfo[2]{#2}
\providecommand\natexlab[1]{#1}
\providecommand\showeprint[2][]{arXiv:#2}

\bibitem[Brown et~al\mbox{.}(2020)]%
        {brown2020language}
\bibfield{author}{\bibinfo{person}{Tom Brown}, \bibinfo{person}{Benjamin Mann}, \bibinfo{person}{Nick Ryder}, \bibinfo{person}{Melanie Subbiah}, \bibinfo{person}{Jared~D Kaplan}, \bibinfo{person}{Prafulla Dhariwal}, \bibinfo{person}{Arvind Neelakantan}, \bibinfo{person}{Pranav Shyam}, \bibinfo{person}{Girish Sastry}, \bibinfo{person}{Amanda Askell}, {et~al\mbox{.}}} \bibinfo{year}{2020}\natexlab{}.
\newblock \showarticletitle{Language models are few-shot learners}.
\newblock \bibinfo{journal}{\emph{Advances in neural information processing systems}}  \bibinfo{volume}{33} (\bibinfo{year}{2020}), \bibinfo{pages}{1877--1901}.
\newblock


\bibitem[Cao et~al\mbox{.}(2021)]%
        {cao2021bipartite}
\bibfield{author}{\bibinfo{person}{Jiangxia Cao}, \bibinfo{person}{Xixun Lin}, \bibinfo{person}{Shu Guo}, \bibinfo{person}{Luchen Liu}, \bibinfo{person}{Tingwen Liu}, {and} \bibinfo{person}{Bin Wang}.} \bibinfo{year}{2021}\natexlab{}.
\newblock \showarticletitle{Bipartite graph embedding via mutual information maximization}. In \bibinfo{booktitle}{\emph{Proceedings of the 14th ACM international conference on web search and data mining}}. \bibinfo{pages}{635--643}.
\newblock


\bibitem[Chen et~al\mbox{.}(2020)]%
        {chen2020graph}
\bibfield{author}{\bibinfo{person}{Liqun Chen}, \bibinfo{person}{Zhe Gan}, \bibinfo{person}{Yu Cheng}, \bibinfo{person}{Linjie Li}, \bibinfo{person}{Lawrence Carin}, {and} \bibinfo{person}{Jingjing Liu}.} \bibinfo{year}{2020}\natexlab{}.
\newblock \showarticletitle{Graph optimal transport for cross-domain alignment}. In \bibinfo{booktitle}{\emph{International Conference on Machine Learning}}. PMLR, \bibinfo{pages}{1542--1553}.
\newblock


\bibitem[Chen et~al\mbox{.}(2023a)]%
        {chen2023ultra}
\bibfield{author}{\bibinfo{person}{Mouxiang Chen}, \bibinfo{person}{Zemin Liu}, \bibinfo{person}{Chenghao Liu}, \bibinfo{person}{Jundong Li}, \bibinfo{person}{Qiheng Mao}, {and} \bibinfo{person}{Jianling Sun}.} \bibinfo{year}{2023}\natexlab{a}.
\newblock \showarticletitle{Ultra-dp: Unifying graph pre-training with multi-task graph dual prompt}.
\newblock \bibinfo{journal}{\emph{arXiv preprint arXiv:2310.14845}} (\bibinfo{year}{2023}).
\newblock


\bibitem[Chen et~al\mbox{.}(2023b)]%
        {chen2023feature}
\bibfield{author}{\bibinfo{person}{Yang Chen}, \bibinfo{person}{Zhonglin Ye}, \bibinfo{person}{Haixing Zhao}, {and} \bibinfo{person}{Ying Wang}.} \bibinfo{year}{2023}\natexlab{b}.
\newblock \showarticletitle{Feature-Based Graph Backdoor Attack in the Node Classification Task}.
\newblock \bibinfo{journal}{\emph{International Journal of Intelligent Systems}} \bibinfo{volume}{2023}, \bibinfo{number}{1} (\bibinfo{year}{2023}), \bibinfo{pages}{5418398}.
\newblock


\bibitem[Cui et~al\mbox{.}(2024)]%
        {cui2025prompt}
\bibfield{author}{\bibinfo{person}{Yuanning Cui}, \bibinfo{person}{Zequn Sun}, {and} \bibinfo{person}{Wei Hu}.} \bibinfo{year}{2024}\natexlab{}.
\newblock \showarticletitle{A Prompt-Based Knowledge Graph Foundation Model for Universal In-Context Reasoning}.
\newblock \bibinfo{journal}{\emph{NeurIPS}}  \bibinfo{volume}{37} (\bibinfo{year}{2024}), \bibinfo{pages}{7095--7124}.
\newblock


\bibitem[Diao et~al\mbox{.}(2022)]%
        {diao2022molcpt}
\bibfield{author}{\bibinfo{person}{Cameron Diao}, \bibinfo{person}{Kaixiong Zhou}, \bibinfo{person}{Zirui Liu}, \bibinfo{person}{Xiao Huang}, {and} \bibinfo{person}{Xia Hu}.} \bibinfo{year}{2022}\natexlab{}.
\newblock \showarticletitle{Molcpt: Molecule continuous prompt tuning to generalize molecular representation learning}.
\newblock \bibinfo{journal}{\emph{arXiv preprint arXiv:2212.10614}} (\bibinfo{year}{2022}).
\newblock


\bibitem[Duan et~al\mbox{.}(2024)]%
        {duan2024g}
\bibfield{author}{\bibinfo{person}{Yutai Duan}, \bibinfo{person}{Jie Liu}, \bibinfo{person}{Shaowei Chen}, \bibinfo{person}{Liyi Chen}, {and} \bibinfo{person}{Jianhua Wu}.} \bibinfo{year}{2024}\natexlab{}.
\newblock \showarticletitle{G-Prompt: Graphon-based Prompt Tuning for graph classification}.
\newblock \bibinfo{journal}{\emph{Information Processing \& Management}} \bibinfo{volume}{61}, \bibinfo{number}{3} (\bibinfo{year}{2024}), \bibinfo{pages}{103639}.
\newblock


\bibitem[Fang et~al\mbox{.}(2023a)]%
        {fang2023gpf}
\bibfield{author}{\bibinfo{person}{Taoran Fang}, \bibinfo{person}{Yunchao Zhang}, \bibinfo{person}{Yang Yang}, \bibinfo{person}{Chunping Wang}, {and} \bibinfo{person}{Lei Chen}.} \bibinfo{year}{2023}\natexlab{a}.
\newblock \showarticletitle{Universal prompt tuning for graph neural networks}.
\newblock \bibinfo{journal}{\emph{Advances in Neural Information Processing Systems}} (\bibinfo{year}{2023}).
\newblock


\bibitem[Fang et~al\mbox{.}(2023b)]%
        {fang2023universal}
\bibfield{author}{\bibinfo{person}{Taoran Fang}, \bibinfo{person}{Yunchao Zhang}, \bibinfo{person}{Yang Yang}, \bibinfo{person}{Chunping Wang}, {and} \bibinfo{person}{Lei Chen}.} \bibinfo{year}{2023}\natexlab{b}.
\newblock \showarticletitle{Universal prompt tuning for graph neural networks}.
\newblock \bibinfo{journal}{\emph{Advances in Neural Information Processing Systems}}  \bibinfo{volume}{36} (\bibinfo{year}{2023}), \bibinfo{pages}{52464--52489}.
\newblock


\bibitem[Fu et~al\mbox{.}(2023)]%
        {fu2023spatial}
\bibfield{author}{\bibinfo{person}{Xingbo Fu}, \bibinfo{person}{Chen Chen}, \bibinfo{person}{Yushun Dong}, \bibinfo{person}{Anil Vullikanti}, \bibinfo{person}{Eili Klein}, \bibinfo{person}{Gregory Madden}, {and} \bibinfo{person}{Jundong Li}.} \bibinfo{year}{2023}\natexlab{}.
\newblock \showarticletitle{Spatial-Temporal Networks for Antibiogram Pattern Prediction}. In \bibinfo{booktitle}{\emph{2023 IEEE 11th International Conference on Healthcare Informatics (ICHI)}}.
\newblock


\bibitem[Fu et~al\mbox{.}(2025)]%
        {fu2025edge}
\bibfield{author}{\bibinfo{person}{Xingbo Fu}, \bibinfo{person}{Yinhan He}, {and} \bibinfo{person}{Jundong Li}.} \bibinfo{year}{2025}\natexlab{}.
\newblock \showarticletitle{Edge Prompt Tuning for Graph Neural Networks}. In \bibinfo{booktitle}{\emph{The Thirteenth International Conference on Learning Representations}}.
\newblock


\bibitem[Gasteiger et~al\mbox{.}(2019)]%
        {gasteiger2019diffusion}
\bibfield{author}{\bibinfo{person}{Johannes Gasteiger}, \bibinfo{person}{Stefan Wei{\ss}enberger}, {and} \bibinfo{person}{Stephan G{\"u}nnemann}.} \bibinfo{year}{2019}\natexlab{}.
\newblock \showarticletitle{Diffusion improves graph learning}.
\newblock \bibinfo{journal}{\emph{Advances in neural information processing systems}} (\bibinfo{year}{2019}).
\newblock


\bibitem[Ge et~al\mbox{.}(2024)]%
        {ge2024psp}
\bibfield{author}{\bibinfo{person}{Qingqing Ge}, \bibinfo{person}{Zeyuan Zhao}, \bibinfo{person}{Yiding Liu}, \bibinfo{person}{Anfeng Cheng}, \bibinfo{person}{Xiang Li}, \bibinfo{person}{Shuaiqiang Wang}, {and} \bibinfo{person}{Dawei Yin}.} \bibinfo{year}{2024}\natexlab{}.
\newblock \showarticletitle{PSP: Pre-training and Structure Prompt Tuning for Graph Neural Networks}. In \bibinfo{booktitle}{\emph{Joint European Conference on Machine Learning and Knowledge Discovery in Databases}}. Springer, \bibinfo{pages}{423--439}.
\newblock


\bibitem[Geng et~al\mbox{.}(2022)]%
        {geng2022recommendation}
\bibfield{author}{\bibinfo{person}{Shijie Geng}, \bibinfo{person}{Shuchang Liu}, \bibinfo{person}{Zuohui Fu}, \bibinfo{person}{Yingqiang Ge}, {and} \bibinfo{person}{Yongfeng Zhang}.} \bibinfo{year}{2022}\natexlab{}.
\newblock \showarticletitle{Recommendation as language processing (RLP): A unified pretrain, personalized prompt \& predict paradigm (P5)}. In \bibinfo{booktitle}{\emph{RecSys}}. \bibinfo{pages}{299--315}.
\newblock


\bibitem[Gong et~al\mbox{.}(2023)]%
        {gong2023prompt}
\bibfield{author}{\bibinfo{person}{Chenghua Gong}, \bibinfo{person}{Xiang Li}, \bibinfo{person}{Jianxiang Yu}, \bibinfo{person}{Cheng Yao}, \bibinfo{person}{Jiaqi Tan}, \bibinfo{person}{Chengcheng Yu}, {and} \bibinfo{person}{Dawei Yin}.} \bibinfo{year}{2023}\natexlab{}.
\newblock \showarticletitle{Prompt tuning for multi-view graph contrastive learning}.
\newblock \bibinfo{journal}{\emph{arXiv preprint arXiv:2310.10362}} (\bibinfo{year}{2023}).
\newblock


\bibitem[Grover and Leskovec(2016)]%
        {grover2016node2vec}
\bibfield{author}{\bibinfo{person}{Aditya Grover} {and} \bibinfo{person}{Jure Leskovec}.} \bibinfo{year}{2016}\natexlab{}.
\newblock \showarticletitle{node2vec: Scalable feature learning for networks}. In \bibinfo{booktitle}{\emph{Proceedings of the 22nd ACM SIGKDD international conference on Knowledge discovery and data mining}}. \bibinfo{pages}{855--864}.
\newblock


\bibitem[Gutmann and Hyv{\"a}rinen(2010)]%
        {gutmann2010noise}
\bibfield{author}{\bibinfo{person}{Michael Gutmann} {and} \bibinfo{person}{Aapo Hyv{\"a}rinen}.} \bibinfo{year}{2010}\natexlab{}.
\newblock \showarticletitle{Noise-contrastive estimation: A new estimation principle for unnormalized statistical models}. In \bibinfo{booktitle}{\emph{Proceedings of the thirteenth international conference on artificial intelligence and statistics}}. JMLR Workshop and Conference Proceedings, \bibinfo{pages}{297--304}.
\newblock


\bibitem[Hamilton et~al\mbox{.}(2017a)]%
        {hamilton2017graphsage}
\bibfield{author}{\bibinfo{person}{Will Hamilton}, \bibinfo{person}{Zhitao Ying}, {and} \bibinfo{person}{Jure Leskovec}.} \bibinfo{year}{2017}\natexlab{a}.
\newblock \showarticletitle{Inductive representation learning on large graphs}. In \bibinfo{booktitle}{\emph{Advances in neural information processing systems}}.
\newblock


\bibitem[Hamilton et~al\mbox{.}(2017b)]%
        {hamilton2017inductive}
\bibfield{author}{\bibinfo{person}{Will Hamilton}, \bibinfo{person}{Zhitao Ying}, {and} \bibinfo{person}{Jure Leskovec}.} \bibinfo{year}{2017}\natexlab{b}.
\newblock \showarticletitle{Inductive representation learning on large graphs}.
\newblock \bibinfo{journal}{\emph{Advances in neural information processing systems}}  \bibinfo{volume}{30} (\bibinfo{year}{2017}).
\newblock


\bibitem[Hassani and Khasahmadi(2020)]%
        {hassani2020contrastive}
\bibfield{author}{\bibinfo{person}{Kaveh Hassani} {and} \bibinfo{person}{Amir~Hosein Khasahmadi}.} \bibinfo{year}{2020}\natexlab{}.
\newblock \showarticletitle{Contrastive multi-view representation learning on graphs}. In \bibinfo{booktitle}{\emph{International conference on machine learning}}. PMLR, \bibinfo{pages}{4116--4126}.
\newblock


\bibitem[He et~al\mbox{.}(2020)]%
        {he2020lightgcn}
\bibfield{author}{\bibinfo{person}{Xiangnan He}, \bibinfo{person}{Kuan Deng}, \bibinfo{person}{Xiang Wang}, \bibinfo{person}{Yan Li}, \bibinfo{person}{Yongdong Zhang}, {and} \bibinfo{person}{Meng Wang}.} \bibinfo{year}{2020}\natexlab{}.
\newblock \showarticletitle{LightGCN: Simplifying and powering graph convolution network for recommendation}. In \bibinfo{booktitle}{\emph{SIGIR}}. \bibinfo{pages}{639--648}.
\newblock


\bibitem[Hou et~al\mbox{.}(2023)]%
        {hou2023graphmae2}
\bibfield{author}{\bibinfo{person}{Zhenyu Hou}, \bibinfo{person}{Yufei He}, \bibinfo{person}{Yukuo Cen}, \bibinfo{person}{Xiao Liu}, \bibinfo{person}{Yuxiao Dong}, \bibinfo{person}{Evgeny Kharlamov}, {and} \bibinfo{person}{Jie Tang}.} \bibinfo{year}{2023}\natexlab{}.
\newblock \showarticletitle{Graphmae2: A decoding-enhanced masked self-supervised graph learner}. In \bibinfo{booktitle}{\emph{Proceedings of the ACM web conference 2023}}.
\newblock


\bibitem[Hou et~al\mbox{.}(2022)]%
        {hou2022graphmae}
\bibfield{author}{\bibinfo{person}{Zhenyu Hou}, \bibinfo{person}{Xiao Liu}, \bibinfo{person}{Yukuo Cen}, \bibinfo{person}{Yuxiao Dong}, \bibinfo{person}{Hongxia Yang}, \bibinfo{person}{Chunjie Wang}, {and} \bibinfo{person}{Jie Tang}.} \bibinfo{year}{2022}\natexlab{}.
\newblock \showarticletitle{Graphmae: Self-supervised masked graph autoencoders}. In \bibinfo{booktitle}{\emph{Proceedings of the 28th ACM SIGKDD Conference on Knowledge Discovery and Data Mining}}.
\newblock


\bibitem[Hu et~al\mbox{.}(2021)]%
        {hu2021ogb}
\bibfield{author}{\bibinfo{person}{Weihua Hu}, \bibinfo{person}{Matthias Fey}, \bibinfo{person}{Hongyu Ren}, \bibinfo{person}{Maho Nakata}, \bibinfo{person}{Yuxiao Dong}, {and} \bibinfo{person}{Jure Leskovec}.} \bibinfo{year}{2021}\natexlab{}.
\newblock \showarticletitle{Ogb-lsc: A large-scale challenge for machine learning on graphs}.
\newblock \bibinfo{journal}{\emph{arXiv preprint arXiv:2103.09430}} (\bibinfo{year}{2021}).
\newblock


\bibitem[Hu et~al\mbox{.}(2020c)]%
        {hu2020ogb}
\bibfield{author}{\bibinfo{person}{Weihua Hu}, \bibinfo{person}{Matthias Fey}, \bibinfo{person}{Marinka Zitnik}, \bibinfo{person}{Yuxiao Dong}, \bibinfo{person}{Hongyu Ren}, \bibinfo{person}{Bowen Liu}, \bibinfo{person}{Michele Catasta}, {and} \bibinfo{person}{Jure Leskovec}.} \bibinfo{year}{2020}\natexlab{c}.
\newblock \showarticletitle{Open graph benchmark: Datasets for machine learning on graphs}.
\newblock \bibinfo{journal}{\emph{Advances in neural information processing systems}} (\bibinfo{year}{2020}).
\newblock


\bibitem[Hu et~al\mbox{.}(2020d)]%
        {hu2019strategies}
\bibfield{author}{\bibinfo{person}{Weihua Hu}, \bibinfo{person}{Bowen Liu}, \bibinfo{person}{Joseph Gomes}, \bibinfo{person}{Marinka Zitnik}, \bibinfo{person}{Percy Liang}, \bibinfo{person}{Vijay Pande}, {and} \bibinfo{person}{Jure Leskovec}.} \bibinfo{year}{2020}\natexlab{d}.
\newblock \showarticletitle{Strategies for pre-training graph neural networks}. In \bibinfo{booktitle}{\emph{International Conference on Learning Representations}}.
\newblock


\bibitem[Hu et~al\mbox{.}(2020b)]%
        {hu2020gpt}
\bibfield{author}{\bibinfo{person}{Ziniu Hu}, \bibinfo{person}{Yuxiao Dong}, \bibinfo{person}{Kuansan Wang}, \bibinfo{person}{Kai-Wei Chang}, {and} \bibinfo{person}{Yizhou Sun}.} \bibinfo{year}{2020}\natexlab{b}.
\newblock \showarticletitle{Gpt-gnn: Generative pre-training of graph neural networks}. In \bibinfo{booktitle}{\emph{Proceedings of the 26th ACM SIGKDD international conference on knowledge discovery \& data mining}}. \bibinfo{pages}{1857--1867}.
\newblock


\bibitem[Hu et~al\mbox{.}(2020a)]%
        {hu2020heterogeneous}
\bibfield{author}{\bibinfo{person}{Ziniu Hu}, \bibinfo{person}{Yuxiao Dong}, \bibinfo{person}{Kuansan Wang}, {and} \bibinfo{person}{Yizhou Sun}.} \bibinfo{year}{2020}\natexlab{a}.
\newblock \showarticletitle{Heterogeneous graph transformer}. In \bibinfo{booktitle}{\emph{Proceedings of the web conference 2020}}. \bibinfo{pages}{2704--2710}.
\newblock


\bibitem[Hu et~al\mbox{.}(2019)]%
        {hu2019pre}
\bibfield{author}{\bibinfo{person}{Ziniu Hu}, \bibinfo{person}{Changjun Fan}, \bibinfo{person}{Ting Chen}, \bibinfo{person}{Kai-Wei Chang}, {and} \bibinfo{person}{Yizhou Sun}.} \bibinfo{year}{2019}\natexlab{}.
\newblock \showarticletitle{Pre-training graph neural networks for generic structural feature extraction}.
\newblock \bibinfo{journal}{\emph{arXiv preprint arXiv:1905.13728}} (\bibinfo{year}{2019}).
\newblock


\bibitem[Huang et~al\mbox{.}(2024b)]%
        {huang2024measuring}
\bibfield{author}{\bibinfo{person}{Renhong Huang}, \bibinfo{person}{Jiarong Xu}, \bibinfo{person}{Xin Jiang}, \bibinfo{person}{Chenglu Pan}, \bibinfo{person}{Zhiming Yang}, \bibinfo{person}{Chunping Wang}, {and} \bibinfo{person}{Yang Yang}.} \bibinfo{year}{2024}\natexlab{b}.
\newblock \showarticletitle{Measuring Task Similarity and Its Implication in Fine-Tuning Graph Neural Networks}. In \bibinfo{booktitle}{\emph{Proceedings of the AAAI Conference on Artificial Intelligence}}.
\newblock


\bibitem[Huang et~al\mbox{.}(2024a)]%
        {huang2024graph}
\bibfield{author}{\bibinfo{person}{Zhenhua Huang}, \bibinfo{person}{Kunhao Li}, \bibinfo{person}{Shaojie Wang}, \bibinfo{person}{Zhaohong Jia}, \bibinfo{person}{Wentao Zhu}, {and} \bibinfo{person}{Sharad Mehrotra}.} \bibinfo{year}{2024}\natexlab{a}.
\newblock \showarticletitle{Graph Structure Prompt Learning: A Novel Methodology to Improve Performance of Graph Neural Networks}.
\newblock \bibinfo{journal}{\emph{arXiv preprint arXiv:2407.11361}} (\bibinfo{year}{2024}).
\newblock


\bibitem[Hwang et~al\mbox{.}(2020)]%
        {hwang2020self}
\bibfield{author}{\bibinfo{person}{Dasol Hwang}, \bibinfo{person}{Jinyoung Park}, \bibinfo{person}{Sunyoung Kwon}, \bibinfo{person}{KyungMin Kim}, \bibinfo{person}{Jung-Woo Ha}, {and} \bibinfo{person}{Hyunwoo~J Kim}.} \bibinfo{year}{2020}\natexlab{}.
\newblock \showarticletitle{Self-supervised auxiliary learning with meta-paths for heterogeneous graphs}.
\newblock \bibinfo{journal}{\emph{Advances in neural information processing systems}}  \bibinfo{volume}{33} (\bibinfo{year}{2020}), \bibinfo{pages}{10294--10305}.
\newblock


\bibitem[Hwang et~al\mbox{.}(2022)]%
        {hwang2022analysis}
\bibfield{author}{\bibinfo{person}{EunJeong Hwang}, \bibinfo{person}{Veronika Thost}, \bibinfo{person}{Shib~Sankar Dasgupta}, {and} \bibinfo{person}{Tengfei Ma}.} \bibinfo{year}{2022}\natexlab{}.
\newblock \showarticletitle{An analysis of virtual nodes in graph neural networks for link prediction}. In \bibinfo{booktitle}{\emph{The First Learning on Graphs Conference}}.
\newblock


\bibitem[Jia et~al\mbox{.}(2022)]%
        {jia2022vpt}
\bibfield{author}{\bibinfo{person}{Menglin Jia}, \bibinfo{person}{Luming Tang}, \bibinfo{person}{Bor-Chun Chen}, \bibinfo{person}{Claire Cardie}, \bibinfo{person}{Serge Belongie}, \bibinfo{person}{Bharath Hariharan}, {and} \bibinfo{person}{Ser-Nam Lim}.} \bibinfo{year}{2022}\natexlab{}.
\newblock \showarticletitle{Visual prompt tuning}. In \bibinfo{booktitle}{\emph{European Conference on Computer Vision}}.
\newblock


\bibitem[Jiang et~al\mbox{.}(2024)]%
        {jiang2024unified}
\bibfield{author}{\bibinfo{person}{Bo Jiang}, \bibinfo{person}{Hao Wu}, \bibinfo{person}{Ziyan Zhang}, \bibinfo{person}{Beibei Wang}, {and} \bibinfo{person}{Jin Tang}.} \bibinfo{year}{2024}\natexlab{}.
\newblock \showarticletitle{A unified graph selective prompt learning for graph neural networks}.
\newblock \bibinfo{journal}{\emph{arXiv preprint arXiv:2406.10498}} (\bibinfo{year}{2024}).
\newblock


\bibitem[Jiang et~al\mbox{.}(2021)]%
        {jiang2021contrastive}
\bibfield{author}{\bibinfo{person}{Xunqiang Jiang}, \bibinfo{person}{Yuanfu Lu}, \bibinfo{person}{Yuan Fang}, {and} \bibinfo{person}{Chuan Shi}.} \bibinfo{year}{2021}\natexlab{}.
\newblock \showarticletitle{Contrastive pre-training of GNNs on heterogeneous graphs}. In \bibinfo{booktitle}{\emph{Proceedings of the 30th ACM international conference on information \& knowledge management}}.
\newblock


\bibitem[Jiao et~al\mbox{.}(2020)]%
        {jiao2020sub}
\bibfield{author}{\bibinfo{person}{Yizhu Jiao}, \bibinfo{person}{Yun Xiong}, \bibinfo{person}{Jiawei Zhang}, \bibinfo{person}{Yao Zhang}, \bibinfo{person}{Tianqi Zhang}, {and} \bibinfo{person}{Yangyong Zhu}.} \bibinfo{year}{2020}\natexlab{}.
\newblock \showarticletitle{Sub-graph contrast for scalable self-supervised graph representation learning}. In \bibinfo{booktitle}{\emph{2020 IEEE international conference on data mining (ICDM)}}. IEEE, \bibinfo{pages}{222--231}.
\newblock


\bibitem[Jin et~al\mbox{.}(2021)]%
        {jin2021multi}
\bibfield{author}{\bibinfo{person}{Ming Jin}, \bibinfo{person}{Yizhen Zheng}, \bibinfo{person}{Yuan-Fang Li}, \bibinfo{person}{Chen Gong}, \bibinfo{person}{Chuan Zhou}, {and} \bibinfo{person}{Shirui Pan}.} \bibinfo{year}{2021}\natexlab{}.
\newblock \showarticletitle{Multi-scale contrastive siamese networks for self-supervised graph representation learning}. In \bibinfo{booktitle}{\emph{International Joint Conference on Artificial Intelligence 2021}}. Association for the Advancement of Artificial Intelligence (AAAI), \bibinfo{pages}{1477--1483}.
\newblock


\bibitem[Jin et~al\mbox{.}(2020)]%
        {jin2020self}
\bibfield{author}{\bibinfo{person}{Wei Jin}, \bibinfo{person}{Tyler Derr}, \bibinfo{person}{Haochen Liu}, \bibinfo{person}{Yiqi Wang}, \bibinfo{person}{Suhang Wang}, \bibinfo{person}{Zitao Liu}, {and} \bibinfo{person}{Jiliang Tang}.} \bibinfo{year}{2020}\natexlab{}.
\newblock \showarticletitle{Self-supervised learning on graphs: Deep insights and new direction}.
\newblock \bibinfo{journal}{\emph{arXiv preprint arXiv:2006.10141}} (\bibinfo{year}{2020}).
\newblock


\bibitem[Jing et~al\mbox{.}(2021)]%
        {jing2021hdmi}
\bibfield{author}{\bibinfo{person}{Baoyu Jing}, \bibinfo{person}{Chanyoung Park}, {and} \bibinfo{person}{Hanghang Tong}.} \bibinfo{year}{2021}\natexlab{}.
\newblock \showarticletitle{Hdmi: High-order deep multiplex infomax}. In \bibinfo{booktitle}{\emph{Proceedings of the web conference 2021}}. \bibinfo{pages}{2414--2424}.
\newblock


\bibitem[Jovanovi{\'c} et~al\mbox{.}(2021)]%
        {jovanovic2021towards}
\bibfield{author}{\bibinfo{person}{Nikola Jovanovi{\'c}}, \bibinfo{person}{Zhao Meng}, \bibinfo{person}{Lukas Faber}, {and} \bibinfo{person}{Roger Wattenhofer}.} \bibinfo{year}{2021}\natexlab{}.
\newblock \showarticletitle{Towards robust graph contrastive learning}.
\newblock \bibinfo{journal}{\emph{arXiv preprint arXiv:2102.13085}} (\bibinfo{year}{2021}).
\newblock


\bibitem[Kefato and Girdzijauskas(2021)]%
        {kefato2021self}
\bibfield{author}{\bibinfo{person}{Zekarias~T Kefato} {and} \bibinfo{person}{Sarunas Girdzijauskas}.} \bibinfo{year}{2021}\natexlab{}.
\newblock \showarticletitle{Self-supervised graph neural networks without explicit negative sampling}.
\newblock \bibinfo{journal}{\emph{arXiv preprint arXiv:2103.14958}} (\bibinfo{year}{2021}).
\newblock


\bibitem[Khattak et~al\mbox{.}(2023)]%
        {khattak2023maple}
\bibfield{author}{\bibinfo{person}{Muhammad~Uzair Khattak}, \bibinfo{person}{Hanoona Rasheed}, \bibinfo{person}{Muhammad Maaz}, \bibinfo{person}{Salman Khan}, {and} \bibinfo{person}{Fahad~Shahbaz Khan}.} \bibinfo{year}{2023}\natexlab{}.
\newblock \showarticletitle{Maple: Multi-modal prompt learning}. In \bibinfo{booktitle}{\emph{Proceedings of the IEEE/CVF Conference on Computer Vision and Pattern Recognition}}. \bibinfo{pages}{19113--19122}.
\newblock


\bibitem[Kim et~al\mbox{.}(2022)]%
        {kim2022graph}
\bibfield{author}{\bibinfo{person}{Dongki Kim}, \bibinfo{person}{Jinheon Baek}, {and} \bibinfo{person}{Sung~Ju Hwang}.} \bibinfo{year}{2022}\natexlab{}.
\newblock \showarticletitle{Graph self-supervised learning with accurate discrepancy learning}.
\newblock \bibinfo{journal}{\emph{Advances in Neural Information Processing Systems}} (\bibinfo{year}{2022}).
\newblock


\bibitem[Kipf and Welling(2016)]%
        {kipf2016variational}
\bibfield{author}{\bibinfo{person}{Thomas~N Kipf} {and} \bibinfo{person}{Max Welling}.} \bibinfo{year}{2016}\natexlab{}.
\newblock \showarticletitle{Variational graph auto-encoders}.
\newblock \bibinfo{journal}{\emph{arXiv preprint arXiv:1611.07308}} (\bibinfo{year}{2016}).
\newblock


\bibitem[Kipf and Welling(2017)]%
        {kipf2016gcn}
\bibfield{author}{\bibinfo{person}{Thomas~N Kipf} {and} \bibinfo{person}{Max Welling}.} \bibinfo{year}{2017}\natexlab{}.
\newblock \showarticletitle{Semi-supervised classification with graph convolutional networks}. In \bibinfo{booktitle}{\emph{International Conference on Learning Representations}}.
\newblock


\bibitem[Lee et~al\mbox{.}(2024)]%
        {lee2024subgraph}
\bibfield{author}{\bibinfo{person}{Junhyun Lee}, \bibinfo{person}{Wooseong Yang}, {and} \bibinfo{person}{Jaewoo Kang}.} \bibinfo{year}{2024}\natexlab{}.
\newblock \showarticletitle{Subgraph-level universal prompt tuning}.
\newblock \bibinfo{journal}{\emph{arXiv preprint arXiv:2402.10380}} (\bibinfo{year}{2024}).
\newblock


\bibitem[Lester et~al\mbox{.}(2021)]%
        {lester2021power}
\bibfield{author}{\bibinfo{person}{Brian Lester}, \bibinfo{person}{Rami Al-Rfou}, {and} \bibinfo{person}{Noah Constant}.} \bibinfo{year}{2021}\natexlab{}.
\newblock \showarticletitle{The power of scale for parameter-efficient prompt tuning}.
\newblock \bibinfo{journal}{\emph{arXiv preprint arXiv:2104.08691}} (\bibinfo{year}{2021}).
\newblock


\bibitem[Li et~al\mbox{.}(2024a)]%
        {li2024hierarchical}
\bibfield{author}{\bibinfo{person}{Jiangmeng Li}, \bibinfo{person}{Yifan Jin}, \bibinfo{person}{Hang Gao}, \bibinfo{person}{Wenwen Qiang}, \bibinfo{person}{Changwen Zheng}, {and} \bibinfo{person}{Fuchun Sun}.} \bibinfo{year}{2024}\natexlab{a}.
\newblock \showarticletitle{Hierarchical topology isomorphism expertise embedded graph contrastive learning}. In \bibinfo{booktitle}{\emph{Proceedings of the AAAI Conference on Artificial Intelligence}}.
\newblock


\bibitem[Li et~al\mbox{.}(2025a)]%
        {li2025instanceaware}
\bibfield{author}{\bibinfo{person}{Jiazheng Li}, \bibinfo{person}{Jundong Li}, {and} \bibinfo{person}{Chuxu Zhang}.} \bibinfo{year}{2025}\natexlab{a}.
\newblock \showarticletitle{Instance-Aware Graph Prompt Learning}.
\newblock \bibinfo{journal}{\emph{Transactions on Machine Learning Research}} (\bibinfo{year}{2025}).
\newblock


\bibitem[Li et~al\mbox{.}(2024b)]%
        {li2024graph}
\bibfield{author}{\bibinfo{person}{Jia Li}, \bibinfo{person}{Xiangguo Sun}, \bibinfo{person}{Yuhan Li}, \bibinfo{person}{Zhixun Li}, \bibinfo{person}{Hong Cheng}, {and} \bibinfo{person}{Jeffrey~Xu Yu}.} \bibinfo{year}{2024}\natexlab{b}.
\newblock \showarticletitle{Graph intelligence with large language models and prompt learning}. In \bibinfo{booktitle}{\emph{Proceedings of the 30th ACM SIGKDD Conference on Knowledge Discovery and Data Mining}}. \bibinfo{pages}{6545--6554}.
\newblock


\bibitem[Li et~al\mbox{.}(2023)]%
        {li2023s}
\bibfield{author}{\bibinfo{person}{Jintang Li}, \bibinfo{person}{Ruofan Wu}, \bibinfo{person}{Wangbin Sun}, \bibinfo{person}{Liang Chen}, \bibinfo{person}{Sheng Tian}, \bibinfo{person}{Liang Zhu}, \bibinfo{person}{Changhua Meng}, \bibinfo{person}{Zibin Zheng}, {and} \bibinfo{person}{Weiqiang Wang}.} \bibinfo{year}{2023}\natexlab{}.
\newblock \showarticletitle{What's behind the mask: Understanding masked graph modeling for graph autoencoders}. In \bibinfo{booktitle}{\emph{Proceedings of the 29th ACM SIGKDD Conference on Knowledge Discovery and Data Mining}}.
\newblock


\bibitem[Li et~al\mbox{.}(2022)]%
        {li2022geomgcl}
\bibfield{author}{\bibinfo{person}{Shuangli Li}, \bibinfo{person}{Jingbo Zhou}, \bibinfo{person}{Tong Xu}, \bibinfo{person}{Dejing Dou}, {and} \bibinfo{person}{Hui Xiong}.} \bibinfo{year}{2022}\natexlab{}.
\newblock \showarticletitle{Geomgcl: Geometric graph contrastive learning for molecular property prediction}. In \bibinfo{booktitle}{\emph{Proceedings of the AAAI conference on artificial intelligence}}.
\newblock


\bibitem[Li et~al\mbox{.}(2021)]%
        {li2021leveraging}
\bibfield{author}{\bibinfo{person}{Xiang Li}, \bibinfo{person}{Danhao Ding}, \bibinfo{person}{Ben Kao}, \bibinfo{person}{Yizhou Sun}, {and} \bibinfo{person}{Nikos Mamoulis}.} \bibinfo{year}{2021}\natexlab{}.
\newblock \showarticletitle{Leveraging meta-path contexts for classification in heterogeneous information networks}. In \bibinfo{booktitle}{\emph{2021 IEEE 37th International Conference on Data Engineering (ICDE)}}. IEEE, \bibinfo{pages}{912--923}.
\newblock


\bibitem[Li et~al\mbox{.}(2025b)]%
        {li2025fairnessaware}
\bibfield{author}{\bibinfo{person}{Zhengpin Li}, \bibinfo{person}{Minhua Lin}, \bibinfo{person}{Jian Wang}, {and} \bibinfo{person}{Suhang Wang}.} \bibinfo{year}{2025}\natexlab{b}.
\newblock \showarticletitle{Fairness-aware Prompt Tuning for Graph Neural Networks}. In \bibinfo{booktitle}{\emph{THE WEB CONFERENCE 2025}}.
\newblock


\bibitem[Lin et~al\mbox{.}(2024)]%
        {lin2024trojan}
\bibfield{author}{\bibinfo{person}{Minhua Lin}, \bibinfo{person}{Zhiwei Zhang}, \bibinfo{person}{Enyan Dai}, \bibinfo{person}{Zongyu Wu}, \bibinfo{person}{Yilong Wang}, \bibinfo{person}{Xiang Zhang}, {and} \bibinfo{person}{Suhang Wang}.} \bibinfo{year}{2024}\natexlab{}.
\newblock \showarticletitle{Trojan Prompt Attacks on Graph Neural Networks}.
\newblock \bibinfo{journal}{\emph{arXiv preprint arXiv:2410.13974}} (\bibinfo{year}{2024}).
\newblock


\bibitem[Liu et~al\mbox{.}(2023b)]%
        {liu2023pre}
\bibfield{author}{\bibinfo{person}{Peng Liu}, \bibinfo{person}{Lemei Zhang}, {and} \bibinfo{person}{Jon~Atle Gulla}.} \bibinfo{year}{2023}\natexlab{b}.
\newblock \showarticletitle{Pre-train, prompt, and recommendation: A comprehensive survey of language modeling paradigm adaptations in recommender systems}.
\newblock \bibinfo{journal}{\emph{TACL}}  \bibinfo{volume}{11} (\bibinfo{year}{2023}), \bibinfo{pages}{1553--1571}.
\newblock


\bibitem[Liu et~al\mbox{.}(2023a)]%
        {liu2023graphprompt}
\bibfield{author}{\bibinfo{person}{Zemin Liu}, \bibinfo{person}{Xingtong Yu}, \bibinfo{person}{Yuan Fang}, {and} \bibinfo{person}{Xinming Zhang}.} \bibinfo{year}{2023}\natexlab{a}.
\newblock \showarticletitle{Graphprompt: Unifying pre-training and downstream tasks for graph neural networks}. In \bibinfo{booktitle}{\emph{Proceedings of the ACM Web Conference 2023}}.
\newblock


\bibitem[Long et~al\mbox{.}(2024a)]%
        {long2024moat}
\bibfield{author}{\bibinfo{person}{Qingqing Long}, \bibinfo{person}{Yuchen Yan}, \bibinfo{person}{Wentao Cui}, \bibinfo{person}{Wei Ju}, \bibinfo{person}{Zhihong Zhu}, \bibinfo{person}{Yuanchun Zhou}, \bibinfo{person}{Xuezhi Wang}, {and} \bibinfo{person}{Meng Xiao}.} \bibinfo{year}{2024}\natexlab{a}.
\newblock \showarticletitle{MOAT: Graph prompting for 3D molecular graphs}. In \bibinfo{booktitle}{\emph{CIKM}}. \bibinfo{pages}{1586--1596}.
\newblock


\bibitem[Long et~al\mbox{.}(2024b)]%
        {long2024towards}
\bibfield{author}{\bibinfo{person}{Qingqing Long}, \bibinfo{person}{Yuchen Yan}, \bibinfo{person}{Peiyan Zhang}, \bibinfo{person}{Chen Fang}, \bibinfo{person}{Wentao Cui}, \bibinfo{person}{Zhiyuan Ning}, \bibinfo{person}{Meng Xiao}, \bibinfo{person}{Ning Cao}, \bibinfo{person}{Xiao Luo}, \bibinfo{person}{Lingjun Xu}, {et~al\mbox{.}}} \bibinfo{year}{2024}\natexlab{b}.
\newblock \showarticletitle{Towards Graph Prompt Learning: A Survey and Beyond}.
\newblock \bibinfo{journal}{\emph{arXiv preprint arXiv:2408.14520}} (\bibinfo{year}{2024}).
\newblock


\bibitem[Luan et~al\mbox{.}(2023)]%
        {luan2023graph}
\bibfield{author}{\bibinfo{person}{Sitao Luan}, \bibinfo{person}{Chenqing Hua}, \bibinfo{person}{Minkai Xu}, \bibinfo{person}{Qincheng Lu}, \bibinfo{person}{Jiaqi Zhu}, \bibinfo{person}{Xiao-Wen Chang}, \bibinfo{person}{Jie Fu}, \bibinfo{person}{Jure Leskovec}, {and} \bibinfo{person}{Doina Precup}.} \bibinfo{year}{2023}\natexlab{}.
\newblock \showarticletitle{When do graph neural networks help with node classification? investigating the homophily principle on node distinguishability}.
\newblock \bibinfo{journal}{\emph{Advances in Neural Information Processing Systems}} (\bibinfo{year}{2023}).
\newblock


\bibitem[Lyu et~al\mbox{.}(2024)]%
        {lyu2024cross}
\bibfield{author}{\bibinfo{person}{Xiaoting Lyu}, \bibinfo{person}{Yufei Han}, \bibinfo{person}{Wei Wang}, \bibinfo{person}{Hangwei Qian}, \bibinfo{person}{Ivor Tsang}, {and} \bibinfo{person}{Xiangliang Zhang}.} \bibinfo{year}{2024}\natexlab{}.
\newblock \showarticletitle{Cross-context backdoor attacks against graph prompt learning}. In \bibinfo{booktitle}{\emph{Proceedings of the 30th ACM SIGKDD Conference on Knowledge Discovery and Data Mining}}. \bibinfo{pages}{2094--2105}.
\newblock


\bibitem[Ma et~al\mbox{.}(2021)]%
        {ma2021calibrating}
\bibfield{author}{\bibinfo{person}{Kaili Ma}, \bibinfo{person}{Haochen Yang}, \bibinfo{person}{Han Yang}, \bibinfo{person}{Yongqiang Chen}, {and} \bibinfo{person}{James Cheng}.} \bibinfo{year}{2021}\natexlab{}.
\newblock \showarticletitle{Calibrating and improving graph contrastive learning}.
\newblock \bibinfo{journal}{\emph{arXiv preprint arXiv:2101.11525}} (\bibinfo{year}{2021}).
\newblock


\bibitem[Ma et~al\mbox{.}(2024)]%
        {ma2024hetgpt}
\bibfield{author}{\bibinfo{person}{Yihong Ma}, \bibinfo{person}{Ning Yan}, \bibinfo{person}{Jiayu Li}, \bibinfo{person}{Masood Mortazavi}, {and} \bibinfo{person}{Nitesh~V Chawla}.} \bibinfo{year}{2024}\natexlab{}.
\newblock \showarticletitle{Hetgpt: Harnessing the power of prompt tuning in pre-trained heterogeneous graph neural networks}. In \bibinfo{booktitle}{\emph{Proceedings of the ACM on Web Conference 2024}}.
\newblock


\bibitem[Mavromatis and Karypis(2020)]%
        {mavromatis2020graph}
\bibfield{author}{\bibinfo{person}{Costas Mavromatis} {and} \bibinfo{person}{George Karypis}.} \bibinfo{year}{2020}\natexlab{}.
\newblock \showarticletitle{Graph infoclust: Leveraging cluster-level node information for unsupervised graph representation learning}.
\newblock \bibinfo{journal}{\emph{arXiv preprint arXiv:2009.06946}} (\bibinfo{year}{2020}).
\newblock


\bibitem[Navarin et~al\mbox{.}(2018)]%
        {navarin2018pre}
\bibfield{author}{\bibinfo{person}{Nicol{\`o} Navarin}, \bibinfo{person}{Dinh~V Tran}, {and} \bibinfo{person}{Alessandro Sperduti}.} \bibinfo{year}{2018}\natexlab{}.
\newblock \showarticletitle{Pre-training graph neural networks with kernels}.
\newblock \bibinfo{journal}{\emph{arXiv preprint arXiv:1811.06930}} (\bibinfo{year}{2018}).
\newblock


\bibitem[Nowozin et~al\mbox{.}(2016)]%
        {nowozin2016f}
\bibfield{author}{\bibinfo{person}{Sebastian Nowozin}, \bibinfo{person}{Botond Cseke}, {and} \bibinfo{person}{Ryota Tomioka}.} \bibinfo{year}{2016}\natexlab{}.
\newblock \showarticletitle{f-gan: Training generative neural samplers using variational divergence minimization}.
\newblock \bibinfo{journal}{\emph{Advances in neural information processing systems}}  \bibinfo{volume}{29} (\bibinfo{year}{2016}).
\newblock


\bibitem[Opolka et~al\mbox{.}(2019)]%
        {opolka2019spatio}
\bibfield{author}{\bibinfo{person}{Felix~L Opolka}, \bibinfo{person}{Aaron Solomon}, \bibinfo{person}{C{\u{a}}t{\u{a}}lina Cangea}, \bibinfo{person}{Petar Veli{\v{c}}kovi{\'c}}, \bibinfo{person}{Pietro Li{\`o}}, {and} \bibinfo{person}{R~Devon Hjelm}.} \bibinfo{year}{2019}\natexlab{}.
\newblock \showarticletitle{Spatio-temporal deep graph infomax}.
\newblock \bibinfo{journal}{\emph{arXiv preprint arXiv:1904.06316}} (\bibinfo{year}{2019}).
\newblock


\bibitem[Park et~al\mbox{.}(2020)]%
        {park2020unsupervised}
\bibfield{author}{\bibinfo{person}{Chanyoung Park}, \bibinfo{person}{Donghyun Kim}, \bibinfo{person}{Jiawei Han}, {and} \bibinfo{person}{Hwanjo Yu}.} \bibinfo{year}{2020}\natexlab{}.
\newblock \showarticletitle{Unsupervised attributed multiplex network embedding}. In \bibinfo{booktitle}{\emph{Proceedings of the AAAI conference on artificial intelligence}}, Vol.~\bibinfo{volume}{34}. \bibinfo{pages}{5371--5378}.
\newblock


\bibitem[Peng et~al\mbox{.}(2024)]%
        {peng2024mmgpl}
\bibfield{author}{\bibinfo{person}{Liang Peng}, \bibinfo{person}{Songyue Cai}, \bibinfo{person}{Zongqian Wu}, \bibinfo{person}{Huifang Shang}, \bibinfo{person}{Xiaofeng Zhu}, {and} \bibinfo{person}{Xiaoxiao Li}.} \bibinfo{year}{2024}\natexlab{}.
\newblock \showarticletitle{Mmgpl: Multimodal medical data analysis with graph prompt learning}.
\newblock \bibinfo{journal}{\emph{Medical Image Analysis}}  \bibinfo{volume}{97} (\bibinfo{year}{2024}), \bibinfo{pages}{103225}.
\newblock


\bibitem[Peng et~al\mbox{.}(2020a)]%
        {peng2020self}
\bibfield{author}{\bibinfo{person}{Zhen Peng}, \bibinfo{person}{Yixiang Dong}, \bibinfo{person}{Minnan Luo}, \bibinfo{person}{Xiao-Ming Wu}, {and} \bibinfo{person}{Qinghua Zheng}.} \bibinfo{year}{2020}\natexlab{a}.
\newblock \showarticletitle{Self-supervised graph representation learning via global context prediction}.
\newblock \bibinfo{journal}{\emph{arXiv preprint arXiv:2003.01604}} (\bibinfo{year}{2020}).
\newblock


\bibitem[Peng et~al\mbox{.}(2020b)]%
        {peng2020graph}
\bibfield{author}{\bibinfo{person}{Zhen Peng}, \bibinfo{person}{Wenbing Huang}, \bibinfo{person}{Minnan Luo}, \bibinfo{person}{Qinghua Zheng}, \bibinfo{person}{Yu Rong}, \bibinfo{person}{Tingyang Xu}, {and} \bibinfo{person}{Junzhou Huang}.} \bibinfo{year}{2020}\natexlab{b}.
\newblock \showarticletitle{Graph representation learning via graphical mutual information maximization}. In \bibinfo{booktitle}{\emph{Proceedings of The Web Conference 2020}}. \bibinfo{pages}{259--270}.
\newblock


\bibitem[Perozzi et~al\mbox{.}(2014)]%
        {perozzi2014deepwalk}
\bibfield{author}{\bibinfo{person}{Bryan Perozzi}, \bibinfo{person}{Rami Al-Rfou}, {and} \bibinfo{person}{Steven Skiena}.} \bibinfo{year}{2014}\natexlab{}.
\newblock \showarticletitle{Deepwalk: Online learning of social representations}. In \bibinfo{booktitle}{\emph{Proceedings of the 20th ACM SIGKDD international conference on Knowledge discovery and data mining}}. \bibinfo{pages}{701--710}.
\newblock


\bibitem[Qi et~al\mbox{.}(2022)]%
        {qi2022graph}
\bibfield{author}{\bibinfo{person}{Jianzhong Qi}, \bibinfo{person}{Zhuowei Zhao}, \bibinfo{person}{Egemen Tanin}, \bibinfo{person}{Tingru Cui}, \bibinfo{person}{Neema Nassir}, {and} \bibinfo{person}{Majid Sarvi}.} \bibinfo{year}{2022}\natexlab{}.
\newblock \showarticletitle{A graph and attentive multi-path convolutional network for traffic prediction}.
\newblock \bibinfo{journal}{\emph{IEEE Transactions on Knowledge and Data Engineering}} (\bibinfo{year}{2022}).
\newblock


\bibitem[Qiu et~al\mbox{.}(2020)]%
        {qiu2020gcc}
\bibfield{author}{\bibinfo{person}{Jiezhong Qiu}, \bibinfo{person}{Qibin Chen}, \bibinfo{person}{Yuxiao Dong}, \bibinfo{person}{Jing Zhang}, \bibinfo{person}{Hongxia Yang}, \bibinfo{person}{Ming Ding}, \bibinfo{person}{Kuansan Wang}, {and} \bibinfo{person}{Jie Tang}.} \bibinfo{year}{2020}\natexlab{}.
\newblock \showarticletitle{Gcc: Graph contrastive coding for graph neural network pre-training}. In \bibinfo{booktitle}{\emph{Proceedings of the 26th ACM SIGKDD international conference on knowledge discovery \& data mining}}. \bibinfo{pages}{1150--1160}.
\newblock


\bibitem[Ramp{\'a}{\v{s}}ek et~al\mbox{.}(2022)]%
        {rampavsek2022recipe}
\bibfield{author}{\bibinfo{person}{Ladislav Ramp{\'a}{\v{s}}ek}, \bibinfo{person}{Michael Galkin}, \bibinfo{person}{Vijay~Prakash Dwivedi}, \bibinfo{person}{Anh~Tuan Luu}, \bibinfo{person}{Guy Wolf}, {and} \bibinfo{person}{Dominique Beaini}.} \bibinfo{year}{2022}\natexlab{}.
\newblock \showarticletitle{Recipe for a general, powerful, scalable graph transformer}.
\newblock \bibinfo{journal}{\emph{Advances in Neural Information Processing Systems}}  \bibinfo{volume}{35} (\bibinfo{year}{2022}), \bibinfo{pages}{14501--14515}.
\newblock


\bibitem[Ren et~al\mbox{.}(2019)]%
        {ren2019heterogeneous}
\bibfield{author}{\bibinfo{person}{Yuxiang Ren}, \bibinfo{person}{Bo Liu}, \bibinfo{person}{Chao Huang}, \bibinfo{person}{Peng Dai}, \bibinfo{person}{Liefeng Bo}, {and} \bibinfo{person}{Jiawei Zhang}.} \bibinfo{year}{2019}\natexlab{}.
\newblock \showarticletitle{Heterogeneous deep graph infomax}.
\newblock \bibinfo{journal}{\emph{arXiv preprint arXiv:1911.08538}} (\bibinfo{year}{2019}).
\newblock


\bibitem[Rong et~al\mbox{.}(2020)]%
        {rong2020self}
\bibfield{author}{\bibinfo{person}{Yu Rong}, \bibinfo{person}{Yatao Bian}, \bibinfo{person}{Tingyang Xu}, \bibinfo{person}{Weiyang Xie}, \bibinfo{person}{Ying Wei}, \bibinfo{person}{Wenbing Huang}, {and} \bibinfo{person}{Junzhou Huang}.} \bibinfo{year}{2020}\natexlab{}.
\newblock \showarticletitle{Self-supervised graph transformer on large-scale molecular data}.
\newblock \bibinfo{journal}{\emph{Advances in neural information processing systems}} (\bibinfo{year}{2020}).
\newblock


\bibitem[Rossi et~al\mbox{.}(2020)]%
        {rossi2020tgn}
\bibfield{author}{\bibinfo{person}{Emanuele Rossi}, \bibinfo{person}{Ben Chamberlain}, \bibinfo{person}{Fabrizio Frasca}, \bibinfo{person}{Davide Eynard}, \bibinfo{person}{Federico Monti}, {and} \bibinfo{person}{Michael Bronstein}.} \bibinfo{year}{2020}\natexlab{}.
\newblock \showarticletitle{Temporal graph networks for deep learning on dynamic graphs}.
\newblock \bibinfo{journal}{\emph{arXiv preprint arXiv:2006.10637}} (\bibinfo{year}{2020}).
\newblock


\bibitem[Skenderi et~al\mbox{.}(2023)]%
        {skenderi2023graph}
\bibfield{author}{\bibinfo{person}{Geri Skenderi}, \bibinfo{person}{Hang Li}, \bibinfo{person}{Jiliang Tang}, {and} \bibinfo{person}{Marco Cristani}.} \bibinfo{year}{2023}\natexlab{}.
\newblock \showarticletitle{Graph-level representation learning with joint-embedding predictive architectures}.
\newblock \bibinfo{journal}{\emph{arXiv preprint arXiv:2309.16014}} (\bibinfo{year}{2023}).
\newblock


\bibitem[Sun et~al\mbox{.}(2020a)]%
        {sun2019infograph}
\bibfield{author}{\bibinfo{person}{Fan-Yun Sun}, \bibinfo{person}{Jordan Hoffmann}, \bibinfo{person}{Vikas Verma}, {and} \bibinfo{person}{Jian Tang}.} \bibinfo{year}{2020}\natexlab{a}.
\newblock \showarticletitle{Infograph: Unsupervised and semi-supervised graph-level representation learning via mutual information maximization}. In \bibinfo{booktitle}{\emph{International Conference on Learning Representations}}.
\newblock


\bibitem[Sun et~al\mbox{.}(2020b)]%
        {sun2020multi}
\bibfield{author}{\bibinfo{person}{Ke Sun}, \bibinfo{person}{Zhouchen Lin}, {and} \bibinfo{person}{Zhanxing Zhu}.} \bibinfo{year}{2020}\natexlab{b}.
\newblock \showarticletitle{Multi-stage self-supervised learning for graph convolutional networks on graphs with few labeled nodes}. In \bibinfo{booktitle}{\emph{Proceedings of the AAAI conference on artificial intelligence}}.
\newblock


\bibitem[Sun et~al\mbox{.}(2022)]%
        {sun2022gppt}
\bibfield{author}{\bibinfo{person}{Mingchen Sun}, \bibinfo{person}{Kaixiong Zhou}, \bibinfo{person}{Xin He}, \bibinfo{person}{Ying Wang}, {and} \bibinfo{person}{Xin Wang}.} \bibinfo{year}{2022}\natexlab{}.
\newblock \showarticletitle{Gppt: Graph pre-training and prompt tuning to generalize graph neural networks}. In \bibinfo{booktitle}{\emph{Proceedings of the 28th ACM SIGKDD Conference on Knowledge Discovery and Data Mining}}.
\newblock


\bibitem[Sun et~al\mbox{.}(2021)]%
        {sun2021sugar}
\bibfield{author}{\bibinfo{person}{Qingyun Sun}, \bibinfo{person}{Jianxin Li}, \bibinfo{person}{Hao Peng}, \bibinfo{person}{Jia Wu}, \bibinfo{person}{Yuanxing Ning}, \bibinfo{person}{Philip~S Yu}, {and} \bibinfo{person}{Lifang He}.} \bibinfo{year}{2021}\natexlab{}.
\newblock \showarticletitle{Sugar: Subgraph neural network with reinforcement pooling and self-supervised mutual information mechanism}. In \bibinfo{booktitle}{\emph{Proceedings of the web conference 2021}}. \bibinfo{pages}{2081--2091}.
\newblock


\bibitem[Sun et~al\mbox{.}(2023a)]%
        {sun2023all}
\bibfield{author}{\bibinfo{person}{Xiangguo Sun}, \bibinfo{person}{Hong Cheng}, \bibinfo{person}{Jia Li}, \bibinfo{person}{Bo Liu}, {and} \bibinfo{person}{Jihong Guan}.} \bibinfo{year}{2023}\natexlab{a}.
\newblock \showarticletitle{All in one: Multi-task prompting for graph neural networks}. In \bibinfo{booktitle}{\emph{Proceedings of the 29th ACM SIGKDD Conference on Knowledge Discovery and Data Mining}}.
\newblock


\bibitem[Sun et~al\mbox{.}(2023b)]%
        {sun2023survey}
\bibfield{author}{\bibinfo{person}{Xiangguo Sun}, \bibinfo{person}{Jiawen Zhang}, \bibinfo{person}{Xixi Wu}, \bibinfo{person}{Hong Cheng}, \bibinfo{person}{Yun Xiong}, {and} \bibinfo{person}{Jia Li}.} \bibinfo{year}{2023}\natexlab{b}.
\newblock \showarticletitle{Graph prompt learning: A comprehensive survey and beyond}.
\newblock \bibinfo{journal}{\emph{arXiv preprint arXiv:2311.16534}} (\bibinfo{year}{2023}).
\newblock


\bibitem[Suresh et~al\mbox{.}(2021)]%
        {suresh2021adversarial}
\bibfield{author}{\bibinfo{person}{Susheel Suresh}, \bibinfo{person}{Pan Li}, \bibinfo{person}{Cong Hao}, {and} \bibinfo{person}{Jennifer Neville}.} \bibinfo{year}{2021}\natexlab{}.
\newblock \showarticletitle{Adversarial graph augmentation to improve graph contrastive learning}.
\newblock \bibinfo{journal}{\emph{Advances in Neural Information Processing Systems}}  \bibinfo{volume}{34} (\bibinfo{year}{2021}), \bibinfo{pages}{15920--15933}.
\newblock


\bibitem[Tan et~al\mbox{.}(2023b)]%
        {tan2023s2gae}
\bibfield{author}{\bibinfo{person}{Qiaoyu Tan}, \bibinfo{person}{Ninghao Liu}, \bibinfo{person}{Xiao Huang}, \bibinfo{person}{Soo-Hyun Choi}, \bibinfo{person}{Li Li}, \bibinfo{person}{Rui Chen}, {and} \bibinfo{person}{Xia Hu}.} \bibinfo{year}{2023}\natexlab{b}.
\newblock \showarticletitle{S2gae: Self-supervised graph autoencoders are generalizable learners with graph masking}. In \bibinfo{booktitle}{\emph{Proceedings of the sixteenth ACM international conference on web search and data mining}}.
\newblock


\bibitem[Tan et~al\mbox{.}(2023a)]%
        {tan2023vnt}
\bibfield{author}{\bibinfo{person}{Zhen Tan}, \bibinfo{person}{Ruocheng Guo}, \bibinfo{person}{Kaize Ding}, {and} \bibinfo{person}{Huan Liu}.} \bibinfo{year}{2023}\natexlab{a}.
\newblock \showarticletitle{Virtual node tuning for few-shot node classification}. In \bibinfo{booktitle}{\emph{Proceedings of the 29th ACM SIGKDD Conference on Knowledge Discovery and Data Mining}}.
\newblock


\bibitem[Tang et~al\mbox{.}(2015)]%
        {tang2015line}
\bibfield{author}{\bibinfo{person}{Jian Tang}, \bibinfo{person}{Meng Qu}, \bibinfo{person}{Mingzhe Wang}, \bibinfo{person}{Ming Zhang}, \bibinfo{person}{Jun Yan}, {and} \bibinfo{person}{Qiaozhu Mei}.} \bibinfo{year}{2015}\natexlab{}.
\newblock \showarticletitle{Line: Large-scale information network embedding}. In \bibinfo{booktitle}{\emph{Proceedings of the 24th international conference on world wide web}}. \bibinfo{pages}{1067--1077}.
\newblock


\bibitem[Thakoor et~al\mbox{.}(2021)]%
        {thakoor2021large}
\bibfield{author}{\bibinfo{person}{Shantanu Thakoor}, \bibinfo{person}{Corentin Tallec}, \bibinfo{person}{Mohammad~Gheshlaghi Azar}, \bibinfo{person}{Mehdi Azabou}, \bibinfo{person}{Eva~L Dyer}, \bibinfo{person}{Remi Munos}, \bibinfo{person}{Petar Veli{\v{c}}kovi{\'c}}, {and} \bibinfo{person}{Michal Valko}.} \bibinfo{year}{2021}\natexlab{}.
\newblock \showarticletitle{Large-scale representation learning on graphs via bootstrapping}.
\newblock \bibinfo{journal}{\emph{arXiv preprint arXiv:2102.06514}} (\bibinfo{year}{2021}).
\newblock


\bibitem[Tian et~al\mbox{.}(2024)]%
        {tian2024graph}
\bibfield{author}{\bibinfo{person}{Yijun Tian}, \bibinfo{person}{Huan Song}, \bibinfo{person}{Zichen Wang}, \bibinfo{person}{Haozhu Wang}, \bibinfo{person}{Ziqing Hu}, \bibinfo{person}{Fang Wang}, \bibinfo{person}{Nitesh~V Chawla}, {and} \bibinfo{person}{Panpan Xu}.} \bibinfo{year}{2024}\natexlab{}.
\newblock \showarticletitle{Graph neural prompting with large language models}. In \bibinfo{booktitle}{\emph{Proceedings of the AAAI Conference on Artificial Intelligence}}, Vol.~\bibinfo{volume}{38}. \bibinfo{pages}{19080--19088}.
\newblock


\bibitem[Veli{\v{c}}kovi{\'c} et~al\mbox{.}(2018)]%
        {velivckovic2017gat}
\bibfield{author}{\bibinfo{person}{Petar Veli{\v{c}}kovi{\'c}}, \bibinfo{person}{Guillem Cucurull}, \bibinfo{person}{Arantxa Casanova}, \bibinfo{person}{Adriana Romero}, \bibinfo{person}{Pietro Lio}, {and} \bibinfo{person}{Yoshua Bengio}.} \bibinfo{year}{2018}\natexlab{}.
\newblock \showarticletitle{Graph attention networks}. In \bibinfo{booktitle}{\emph{International Conference on Learning Representations}}.
\newblock


\bibitem[Velickovic et~al\mbox{.}(2019)]%
        {velickovic2019deep}
\bibfield{author}{\bibinfo{person}{Petar Velickovic}, \bibinfo{person}{William Fedus}, \bibinfo{person}{William~L Hamilton}, \bibinfo{person}{Pietro Li{\`o}}, \bibinfo{person}{Yoshua Bengio}, {and} \bibinfo{person}{R~Devon Hjelm}.} \bibinfo{year}{2019}\natexlab{}.
\newblock \showarticletitle{Deep graph infomax.}
\newblock \bibinfo{journal}{\emph{ICLR (poster)}} \bibinfo{volume}{2}, \bibinfo{number}{3} (\bibinfo{year}{2019}), \bibinfo{pages}{4}.
\newblock


\bibitem[Verma et~al\mbox{.}(2021)]%
        {verma2021towards}
\bibfield{author}{\bibinfo{person}{Vikas Verma}, \bibinfo{person}{Thang Luong}, \bibinfo{person}{Kenji Kawaguchi}, \bibinfo{person}{Hieu Pham}, {and} \bibinfo{person}{Quoc Le}.} \bibinfo{year}{2021}\natexlab{}.
\newblock \showarticletitle{Towards domain-agnostic contrastive learning}. In \bibinfo{booktitle}{\emph{International Conference on Machine Learning}}. PMLR, \bibinfo{pages}{10530--10541}.
\newblock


\bibitem[Wan et~al\mbox{.}(2021)]%
        {wan2021contrastive}
\bibfield{author}{\bibinfo{person}{Sheng Wan}, \bibinfo{person}{Yibing Zhan}, \bibinfo{person}{Liu Liu}, \bibinfo{person}{Baosheng Yu}, \bibinfo{person}{Shirui Pan}, {and} \bibinfo{person}{Chen Gong}.} \bibinfo{year}{2021}\natexlab{}.
\newblock \showarticletitle{Contrastive graph poisson networks: Semi-supervised learning with extremely limited labels}.
\newblock \bibinfo{journal}{\emph{Advances in Neural Information Processing Systems}}  \bibinfo{volume}{34} (\bibinfo{year}{2021}), \bibinfo{pages}{6316--6327}.
\newblock


\bibitem[Wang et~al\mbox{.}(2021a)]%
        {wang2021certified}
\bibfield{author}{\bibinfo{person}{Binghui Wang}, \bibinfo{person}{Jinyuan Jia}, \bibinfo{person}{Xiaoyu Cao}, {and} \bibinfo{person}{Neil~Zhenqiang Gong}.} \bibinfo{year}{2021}\natexlab{a}.
\newblock \showarticletitle{Certified robustness of graph neural networks against adversarial structural perturbation}. In \bibinfo{booktitle}{\emph{Proceedings of the 27th ACM SIGKDD Conference on Knowledge Discovery \& Data Mining}}. \bibinfo{pages}{1645--1653}.
\newblock


\bibitem[Wang and Liu(2021)]%
        {wang2021learning}
\bibfield{author}{\bibinfo{person}{Chenguang Wang} {and} \bibinfo{person}{Ziwen Liu}.} \bibinfo{year}{2021}\natexlab{}.
\newblock \showarticletitle{Learning graph representation by aggregating subgraphs via mutual information maximization}.
\newblock \bibinfo{journal}{\emph{arXiv preprint arXiv:2103.13125}} (\bibinfo{year}{2021}).
\newblock


\bibitem[Wang et~al\mbox{.}(2024a)]%
        {wang2024novel}
\bibfield{author}{\bibinfo{person}{Jingchao Wang}, \bibinfo{person}{Zhengnan Deng}, \bibinfo{person}{Tongxu Lin}, \bibinfo{person}{Wenyuan Li}, {and} \bibinfo{person}{Shaobin Ling}.} \bibinfo{year}{2024}\natexlab{a}.
\newblock \showarticletitle{A Novel Prompt Tuning for Graph Transformers: Tailoring Prompts to Graph Topologies}. In \bibinfo{booktitle}{\emph{KDD}}. \bibinfo{pages}{3116--3127}.
\newblock


\bibitem[Wang et~al\mbox{.}(2024b)]%
        {wang2024does}
\bibfield{author}{\bibinfo{person}{Qunzhong Wang}, \bibinfo{person}{Xiangguo Sun}, {and} \bibinfo{person}{Hong Cheng}.} \bibinfo{year}{2024}\natexlab{b}.
\newblock \showarticletitle{Does Graph Prompt Work? A Data Operation Perspective with Theoretical Analysis}.
\newblock \bibinfo{journal}{\emph{arXiv preprint arXiv:2410.01635}} (\bibinfo{year}{2024}).
\newblock


\bibitem[Wang et~al\mbox{.}(2021b)]%
        {wang2021self}
\bibfield{author}{\bibinfo{person}{Xiao Wang}, \bibinfo{person}{Nian Liu}, \bibinfo{person}{Hui Han}, {and} \bibinfo{person}{Chuan Shi}.} \bibinfo{year}{2021}\natexlab{b}.
\newblock \showarticletitle{Self-supervised heterogeneous graph neural network with co-contrastive learning}. In \bibinfo{booktitle}{\emph{Proceedings of the 27th ACM SIGKDD conference on knowledge discovery \& data mining}}. \bibinfo{pages}{1726--1736}.
\newblock


\bibitem[Wang et~al\mbox{.}(2019)]%
        {wang2019dynamic}
\bibfield{author}{\bibinfo{person}{Yue Wang}, \bibinfo{person}{Yongbin Sun}, \bibinfo{person}{Ziwei Liu}, \bibinfo{person}{Sanjay~E Sarma}, \bibinfo{person}{Michael~M Bronstein}, {and} \bibinfo{person}{Justin~M Solomon}.} \bibinfo{year}{2019}\natexlab{}.
\newblock \showarticletitle{Dynamic graph cnn for learning on point clouds}.
\newblock \bibinfo{journal}{\emph{ACM Transactions on Graphics}} (\bibinfo{year}{2019}).
\newblock


\bibitem[Wang et~al\mbox{.}(2024c)]%
        {wang2024ddiprompt}
\bibfield{author}{\bibinfo{person}{Yingying Wang}, \bibinfo{person}{Yun Xiong}, \bibinfo{person}{Xixi Wu}, \bibinfo{person}{Xiangguo Sun}, \bibinfo{person}{Jiawei Zhang}, {and} \bibinfo{person}{GuangYong Zheng}.} \bibinfo{year}{2024}\natexlab{c}.
\newblock \showarticletitle{Ddiprompt: Drug-drug interaction event prediction based on graph prompt learning}. In \bibinfo{booktitle}{\emph{Proceedings of the 33rd ACM International Conference on Information and Knowledge Management}}. \bibinfo{pages}{2431--2441}.
\newblock


\bibitem[Wang et~al\mbox{.}(2025a)]%
        {wang2025graph}
\bibfield{author}{\bibinfo{person}{Zehong Wang}, \bibinfo{person}{Zheyuan Liu}, \bibinfo{person}{Tianyi Ma}, \bibinfo{person}{Jiazheng Li}, \bibinfo{person}{Zheyuan Zhang}, \bibinfo{person}{Xingbo Fu}, \bibinfo{person}{Yiyang Li}, \bibinfo{person}{Zhengqing Yuan}, \bibinfo{person}{Wei Song}, \bibinfo{person}{Yijun Ma}, {et~al\mbox{.}}} \bibinfo{year}{2025}\natexlab{a}.
\newblock \showarticletitle{Graph Foundation Models: A Comprehensive Survey}.
\newblock \bibinfo{journal}{\emph{arXiv preprint arXiv:2505.15116}} (\bibinfo{year}{2025}).
\newblock


\bibitem[Wang et~al\mbox{.}(2024d)]%
        {wang2024gft}
\bibfield{author}{\bibinfo{person}{Zehong Wang}, \bibinfo{person}{Zheyuan Zhang}, \bibinfo{person}{Nitesh~V Chawla}, \bibinfo{person}{Chuxu Zhang}, {and} \bibinfo{person}{Yanfang Ye}.} \bibinfo{year}{2024}\natexlab{d}.
\newblock \showarticletitle{{GFT}: Graph Foundation Model with Transferable Tree Vocabulary}. In \bibinfo{booktitle}{\emph{NeurIPS}}.
\newblock


\bibitem[Wang et~al\mbox{.}(2025b)]%
        {wang2025git}
\bibfield{author}{\bibinfo{person}{Zehong Wang}, \bibinfo{person}{Zheyuan Zhang}, \bibinfo{person}{Tianyi Ma}, \bibinfo{person}{Nitesh~V Chawla}, \bibinfo{person}{Chuxu Zhang}, {and} \bibinfo{person}{Yanfang Ye}.} \bibinfo{year}{2025}\natexlab{b}.
\newblock \showarticletitle{Towards Graph Foundation Models: Learning Generalities Across Graphs via Task-Trees}. In \bibinfo{booktitle}{\emph{ICML}}.
\newblock


\bibitem[Wei et~al\mbox{.}(2024)]%
        {wei2024llmrec}
\bibfield{author}{\bibinfo{person}{Wei Wei}, \bibinfo{person}{Xubin Ren}, \bibinfo{person}{Jiabin Tang}, \bibinfo{person}{Qinyong Wang}, \bibinfo{person}{Lixin Su}, \bibinfo{person}{Suqi Cheng}, \bibinfo{person}{Junfeng Wang}, \bibinfo{person}{Dawei Yin}, {and} \bibinfo{person}{Chao Huang}.} \bibinfo{year}{2024}\natexlab{}.
\newblock \showarticletitle{Llmrec: Large language models with graph augmentation for recommendation}. In \bibinfo{booktitle}{\emph{Proceedings of the 17th ACM International Conference on Web Search and Data Mining}}. \bibinfo{pages}{806--815}.
\newblock


\bibitem[Xia et~al\mbox{.}(2022)]%
        {xia2022simgrace}
\bibfield{author}{\bibinfo{person}{Jun Xia}, \bibinfo{person}{Lirong Wu}, \bibinfo{person}{Jintao Chen}, \bibinfo{person}{Bozhen Hu}, {and} \bibinfo{person}{Stan~Z Li}.} \bibinfo{year}{2022}\natexlab{}.
\newblock \showarticletitle{Simgrace: A simple framework for graph contrastive learning without data augmentation}. In \bibinfo{booktitle}{\emph{Proceedings of the ACM Web Conference 2022}}.
\newblock


\bibitem[Xu et~al\mbox{.}(2024)]%
        {xu2024attacks}
\bibfield{author}{\bibinfo{person}{Ying Xu}, \bibinfo{person}{Michael Lanier}, \bibinfo{person}{Anindya Sarkar}, {and} \bibinfo{person}{Yevgeniy Vorobeychik}.} \bibinfo{year}{2024}\natexlab{}.
\newblock \showarticletitle{Attacks on node attributes in graph neural networks}.
\newblock \bibinfo{journal}{\emph{arXiv preprint arXiv:2402.12426}} (\bibinfo{year}{2024}).
\newblock


\bibitem[Yang et~al\mbox{.}(2020)]%
        {yang2020graph}
\bibfield{author}{\bibinfo{person}{Fang Yang}, \bibinfo{person}{Kunjie Fan}, \bibinfo{person}{Dandan Song}, {and} \bibinfo{person}{Huakang Lin}.} \bibinfo{year}{2020}\natexlab{}.
\newblock \showarticletitle{Graph-based prediction of protein-protein interactions with attributed signed graph embedding}.
\newblock \bibinfo{journal}{\emph{BMC bioinformatics}} (\bibinfo{year}{2020}).
\newblock


\bibitem[Yang et~al\mbox{.}(2023)]%
        {yang2023empirical}
\bibfield{author}{\bibinfo{person}{Haoran Yang}, \bibinfo{person}{Xiangyu Zhao}, \bibinfo{person}{Yicong Li}, \bibinfo{person}{Hongxu Chen}, {and} \bibinfo{person}{Guandong Xu}.} \bibinfo{year}{2023}\natexlab{}.
\newblock \showarticletitle{An empirical study towards prompt-tuning for graph contrastive pre-training in recommendations}.
\newblock \bibinfo{journal}{\emph{NeurIPS}}  \bibinfo{volume}{36} (\bibinfo{year}{2023}), \bibinfo{pages}{62853--62868}.
\newblock


\bibitem[Yang et~al\mbox{.}(2024)]%
        {yang2024graphpro}
\bibfield{author}{\bibinfo{person}{Yuhao Yang}, \bibinfo{person}{Lianghao Xia}, \bibinfo{person}{Da Luo}, \bibinfo{person}{Kangyi Lin}, {and} \bibinfo{person}{Chao Huang}.} \bibinfo{year}{2024}\natexlab{}.
\newblock \showarticletitle{Graphpro: Graph pre-training and prompt learning for recommendation}. In \bibinfo{booktitle}{\emph{WWW}}. \bibinfo{pages}{3690--3699}.
\newblock


\bibitem[Ye et~al\mbox{.}(2023)]%
        {ye2023language}
\bibfield{author}{\bibinfo{person}{Ruosong Ye}, \bibinfo{person}{Caiqi Zhang}, \bibinfo{person}{Runhui Wang}, \bibinfo{person}{Shuyuan Xu}, {and} \bibinfo{person}{Yongfeng Zhang}.} \bibinfo{year}{2023}\natexlab{}.
\newblock \showarticletitle{Language is all a graph needs}.
\newblock \bibinfo{journal}{\emph{arXiv preprint arXiv:2308.07134}} (\bibinfo{year}{2023}).
\newblock


\bibitem[Yi et~al\mbox{.}(2023)]%
        {yi2023contrastive}
\bibfield{author}{\bibinfo{person}{Zixuan Yi}, \bibinfo{person}{Iadh Ounis}, {and} \bibinfo{person}{Craig Macdonald}.} \bibinfo{year}{2023}\natexlab{}.
\newblock \showarticletitle{Contrastive graph prompt-tuning for cross-domain recommendation}.
\newblock \bibinfo{journal}{\emph{ACM Transactions on Information Systems}} \bibinfo{volume}{42}, \bibinfo{number}{2} (\bibinfo{year}{2023}), \bibinfo{pages}{1--28}.
\newblock


\bibitem[Yoo et~al\mbox{.}(2023)]%
        {yoo2023improving}
\bibfield{author}{\bibinfo{person}{Seungryong Yoo}, \bibinfo{person}{Eunji Kim}, \bibinfo{person}{Dahuin Jung}, \bibinfo{person}{Jungbeom Lee}, {and} \bibinfo{person}{Sungroh Yoon}.} \bibinfo{year}{2023}\natexlab{}.
\newblock \showarticletitle{Improving visual prompt tuning for self-supervised vision transformers}. In \bibinfo{booktitle}{\emph{International Conference on Machine Learning}}.
\newblock


\bibitem[You et~al\mbox{.}(2021)]%
        {you2021graph}
\bibfield{author}{\bibinfo{person}{Yuning You}, \bibinfo{person}{Tianlong Chen}, \bibinfo{person}{Yang Shen}, {and} \bibinfo{person}{Zhangyang Wang}.} \bibinfo{year}{2021}\natexlab{}.
\newblock \showarticletitle{Graph contrastive learning automated}. In \bibinfo{booktitle}{\emph{International conference on machine learning}}. PMLR, \bibinfo{pages}{12121--12132}.
\newblock


\bibitem[You et~al\mbox{.}(2020a)]%
        {you2020graphcl}
\bibfield{author}{\bibinfo{person}{Yuning You}, \bibinfo{person}{Tianlong Chen}, \bibinfo{person}{Yongduo Sui}, \bibinfo{person}{Ting Chen}, \bibinfo{person}{Zhangyang Wang}, {and} \bibinfo{person}{Yang Shen}.} \bibinfo{year}{2020}\natexlab{a}.
\newblock \showarticletitle{Graph contrastive learning with augmentations}.
\newblock \bibinfo{journal}{\emph{Advances in neural information processing systems}} (\bibinfo{year}{2020}).
\newblock


\bibitem[You et~al\mbox{.}(2020b)]%
        {you2020graph}
\bibfield{author}{\bibinfo{person}{Yuning You}, \bibinfo{person}{Tianlong Chen}, \bibinfo{person}{Yongduo Sui}, \bibinfo{person}{Ting Chen}, \bibinfo{person}{Zhangyang Wang}, {and} \bibinfo{person}{Yang Shen}.} \bibinfo{year}{2020}\natexlab{b}.
\newblock \showarticletitle{Graph contrastive learning with augmentations}.
\newblock \bibinfo{journal}{\emph{Advances in neural information processing systems}}  \bibinfo{volume}{33} (\bibinfo{year}{2020}), \bibinfo{pages}{5812--5823}.
\newblock


\bibitem[You et~al\mbox{.}(2020c)]%
        {you2020does}
\bibfield{author}{\bibinfo{person}{Yuning You}, \bibinfo{person}{Tianlong Chen}, \bibinfo{person}{Zhangyang Wang}, {and} \bibinfo{person}{Yang Shen}.} \bibinfo{year}{2020}\natexlab{c}.
\newblock \showarticletitle{When does self-supervision help graph convolutional networks?}. In \bibinfo{booktitle}{\emph{international conference on machine learning}}.
\newblock


\bibitem[Yu et~al\mbox{.}(2021)]%
        {yu2021socially}
\bibfield{author}{\bibinfo{person}{Junliang Yu}, \bibinfo{person}{Hongzhi Yin}, \bibinfo{person}{Min Gao}, \bibinfo{person}{Xin Xia}, \bibinfo{person}{Xiangliang Zhang}, {and} \bibinfo{person}{Nguyen~Quoc Viet~Hung}.} \bibinfo{year}{2021}\natexlab{}.
\newblock \showarticletitle{Socially-aware self-supervised tri-training for recommendation}. In \bibinfo{booktitle}{\emph{Proceedings of the 27th ACM SIGKDD conference on knowledge discovery \& data mining}}. \bibinfo{pages}{2084--2092}.
\newblock


\bibitem[Yu et~al\mbox{.}(2024a)]%
        {yu2024generalized}
\bibfield{author}{\bibinfo{person}{Xingtong Yu}, \bibinfo{person}{Zhenghao Liu}, \bibinfo{person}{Yuan Fang}, \bibinfo{person}{Zemin Liu}, \bibinfo{person}{Sihong Chen}, {and} \bibinfo{person}{Xinming Zhang}.} \bibinfo{year}{2024}\natexlab{a}.
\newblock \showarticletitle{Generalized graph prompt: Toward a unification of pre-training and downstream tasks on graphs}.
\newblock \bibinfo{journal}{\emph{IEEE Transactions on Knowledge and Data Engineering}} (\bibinfo{year}{2024}).
\newblock


\bibitem[Yu et~al\mbox{.}(2025a)]%
        {yu2025nodetime}
\bibfield{author}{\bibinfo{person}{Xingtong Yu}, \bibinfo{person}{Zhenghao Liu}, \bibinfo{person}{Xinming Zhang}, {and} \bibinfo{person}{Yuan Fang}.} \bibinfo{year}{2025}\natexlab{a}.
\newblock \showarticletitle{Node-Time Conditional Prompt Learning in Dynamic Graphs}. In \bibinfo{booktitle}{\emph{The Thirteenth International Conference on Learning Representations}}.
\newblock
\urldef\tempurl%
\url{https://openreview.net/forum?id=kVlfYvIqaK}
\showURL{%
\tempurl}


\bibitem[Yu et~al\mbox{.}(2024b)]%
        {yu2024non}
\bibfield{author}{\bibinfo{person}{Xingtong Yu}, \bibinfo{person}{Jie Zhang}, \bibinfo{person}{Yuan Fang}, {and} \bibinfo{person}{Renhe Jiang}.} \bibinfo{year}{2024}\natexlab{b}.
\newblock \showarticletitle{Non-homophilic graph pre-training and prompt learning}.
\newblock \bibinfo{journal}{\emph{arXiv preprint arXiv:2408.12594}} (\bibinfo{year}{2024}).
\newblock


\bibitem[Yu et~al\mbox{.}(2025b)]%
        {yu2025pronog}
\bibfield{author}{\bibinfo{person}{Xingtong Yu}, \bibinfo{person}{Jie Zhang}, \bibinfo{person}{Yuan Fang}, {and} \bibinfo{person}{Renhe Jiang}.} \bibinfo{year}{2025}\natexlab{b}.
\newblock \showarticletitle{Non-homophilic graph pre-training and prompt learning}. In \bibinfo{booktitle}{\emph{Proceedings of the 31st ACM SIGKDD Conference on Knowledge Discovery and Data Mining}}.
\newblock


\bibitem[Yu et~al\mbox{.}(2024c)]%
        {yu2024multigprompt}
\bibfield{author}{\bibinfo{person}{Xingtong Yu}, \bibinfo{person}{Chang Zhou}, \bibinfo{person}{Yuan Fang}, {and} \bibinfo{person}{Xinming Zhang}.} \bibinfo{year}{2024}\natexlab{c}.
\newblock \showarticletitle{MultiGPrompt for multi-task pre-training and prompting on graphs}. In \bibinfo{booktitle}{\emph{Proceedings of the ACM on Web Conference 2024}}.
\newblock


\bibitem[Yun et~al\mbox{.}(2019)]%
        {yun2019graph}
\bibfield{author}{\bibinfo{person}{Seongjun Yun}, \bibinfo{person}{Minbyul Jeong}, \bibinfo{person}{Raehyun Kim}, \bibinfo{person}{Jaewoo Kang}, {and} \bibinfo{person}{Hyunwoo~J Kim}.} \bibinfo{year}{2019}\natexlab{}.
\newblock \showarticletitle{Graph transformer networks}.
\newblock \bibinfo{journal}{\emph{Advances in neural information processing systems}}  \bibinfo{volume}{32} (\bibinfo{year}{2019}).
\newblock


\bibitem[Zeng and Xie(2021)]%
        {zeng2021contrastive}
\bibfield{author}{\bibinfo{person}{Jiaqi Zeng} {and} \bibinfo{person}{Pengtao Xie}.} \bibinfo{year}{2021}\natexlab{}.
\newblock \showarticletitle{Contrastive self-supervised learning for graph classification}. In \bibinfo{booktitle}{\emph{Proceedings of the AAAI conference on Artificial Intelligence}}, Vol.~\bibinfo{volume}{35}. \bibinfo{pages}{10824--10832}.
\newblock


\bibitem[Zhang et~al\mbox{.}(2023a)]%
        {zhang2023iterative}
\bibfield{author}{\bibinfo{person}{Hanlin Zhang}, \bibinfo{person}{Shuai Lin}, \bibinfo{person}{Weiyang Liu}, \bibinfo{person}{Pan Zhou}, \bibinfo{person}{Jian Tang}, \bibinfo{person}{Xiaodan Liang}, {and} \bibinfo{person}{Eric~P Xing}.} \bibinfo{year}{2023}\natexlab{a}.
\newblock \showarticletitle{Iterative graph self-distillation}.
\newblock \bibinfo{journal}{\emph{IEEE Transactions on Knowledge and Data Engineering}} \bibinfo{volume}{36}, \bibinfo{number}{3} (\bibinfo{year}{2023}), \bibinfo{pages}{1161--1169}.
\newblock


\bibitem[Zhang et~al\mbox{.}(2024c)]%
        {zhang2024gpt4rec}
\bibfield{author}{\bibinfo{person}{Peiyan Zhang}, \bibinfo{person}{Yuchen Yan}, \bibinfo{person}{Xi Zhang}, \bibinfo{person}{Liying Kang}, \bibinfo{person}{Chaozhuo Li}, \bibinfo{person}{Feiran Huang}, \bibinfo{person}{Senzhang Wang}, {and} \bibinfo{person}{Sunghun Kim}.} \bibinfo{year}{2024}\natexlab{c}.
\newblock \showarticletitle{Gpt4rec: Graph prompt tuning for streaming recommendation}. In \bibinfo{booktitle}{\emph{Proceedings of the 47th International ACM SIGIR Conference on Research and Development in Information Retrieval}}. \bibinfo{pages}{1774--1784}.
\newblock


\bibitem[Zhang et~al\mbox{.}(2024b)]%
        {zhang2024motif}
\bibfield{author}{\bibinfo{person}{Shichang Zhang}, \bibinfo{person}{Ziniu Hu}, \bibinfo{person}{Arjun Subramonian}, {and} \bibinfo{person}{Yizhou Sun}.} \bibinfo{year}{2024}\natexlab{b}.
\newblock \showarticletitle{Motif-driven contrastive learning of graph representations}.
\newblock \bibinfo{journal}{\emph{IEEE Transactions on Knowledge and Data Engineering}} \bibinfo{volume}{36}, \bibinfo{number}{8} (\bibinfo{year}{2024}), \bibinfo{pages}{4063--4075}.
\newblock


\bibitem[Zhang et~al\mbox{.}(2023b)]%
        {zhang2023structure}
\bibfield{author}{\bibinfo{person}{Wen Zhang}, \bibinfo{person}{Yushan Zhu}, \bibinfo{person}{Mingyang Chen}, \bibinfo{person}{Yuxia Geng}, \bibinfo{person}{Yufeng Huang}, \bibinfo{person}{Yajing Xu}, \bibinfo{person}{Wenting Song}, {and} \bibinfo{person}{Huajun Chen}.} \bibinfo{year}{2023}\natexlab{b}.
\newblock \showarticletitle{Structure pretraining and prompt tuning for knowledge graph transfer}. In \bibinfo{booktitle}{\emph{WWW}}. \bibinfo{pages}{2581--2590}.
\newblock


\bibitem[Zhang et~al\mbox{.}(2024a)]%
        {zhang2024multi}
\bibfield{author}{\bibinfo{person}{Yichi Zhang}, \bibinfo{person}{Binbin Hu}, \bibinfo{person}{Zhuo Chen}, \bibinfo{person}{Lingbing Guo}, \bibinfo{person}{Ziqi Liu}, \bibinfo{person}{Zhiqiang Zhang}, \bibinfo{person}{Lei Liang}, \bibinfo{person}{Huajun Chen}, {and} \bibinfo{person}{Wen Zhang}.} \bibinfo{year}{2024}\natexlab{a}.
\newblock \showarticletitle{Multi-domain Knowledge Graph Collaborative Pre-training and Prompt Tuning for Diverse Downstream Tasks}.
\newblock \bibinfo{journal}{\emph{arXiv preprint arXiv:2405.13085}} (\bibinfo{year}{2024}).
\newblock


\bibitem[Zhao et~al\mbox{.}(2023)]%
        {zhao2023cross}
\bibfield{author}{\bibinfo{person}{Chuang Zhao}, \bibinfo{person}{Hongke Zhao}, \bibinfo{person}{Xiaomeng Li}, \bibinfo{person}{Ming He}, \bibinfo{person}{Jiahui Wang}, {and} \bibinfo{person}{Jianping Fan}.} \bibinfo{year}{2023}\natexlab{}.
\newblock \showarticletitle{Cross-domain recommendation via progressive structural alignment}.
\newblock \bibinfo{journal}{\emph{IEEE Transactions on Knowledge and Data Engineering}} \bibinfo{volume}{36}, \bibinfo{number}{6} (\bibinfo{year}{2023}), \bibinfo{pages}{2401--2415}.
\newblock


\bibitem[Zhao et~al\mbox{.}(2024)]%
        {zhao2024imbalanced}
\bibfield{author}{\bibinfo{person}{Tianxiang Zhao}, \bibinfo{person}{Xiang Zhang}, {and} \bibinfo{person}{Suhang Wang}.} \bibinfo{year}{2024}\natexlab{}.
\newblock \showarticletitle{Imbalanced node classification with synthetic over-sampling}.
\newblock \bibinfo{journal}{\emph{IEEE Transactions on Knowledge and Data Engineering}} (\bibinfo{year}{2024}).
\newblock


\bibitem[Zhili et~al\mbox{.}(2024)]%
        {zhili2024search}
\bibfield{author}{\bibinfo{person}{WANG Zhili}, \bibinfo{person}{DI Shimin}, \bibinfo{person}{CHEN Lei}, {and} \bibinfo{person}{ZHOU Xiaofang}.} \bibinfo{year}{2024}\natexlab{}.
\newblock \showarticletitle{Search to fine-tune pre-trained graph neural networks for graph-level tasks}. In \bibinfo{booktitle}{\emph{2024 IEEE 40th International Conference on Data Engineering (ICDE)}}.
\newblock


\bibitem[Zhou et~al\mbox{.}(2022a)]%
        {zhou2022conditional}
\bibfield{author}{\bibinfo{person}{Kaiyang Zhou}, \bibinfo{person}{Jingkang Yang}, \bibinfo{person}{Chen~Change Loy}, {and} \bibinfo{person}{Ziwei Liu}.} \bibinfo{year}{2022}\natexlab{a}.
\newblock \showarticletitle{Conditional prompt learning for vision-language models}. In \bibinfo{booktitle}{\emph{Proceedings of the IEEE/CVF conference on computer vision and pattern recognition}}.
\newblock


\bibitem[Zhou et~al\mbox{.}(2022b)]%
        {zhou2022coop}
\bibfield{author}{\bibinfo{person}{Kaiyang Zhou}, \bibinfo{person}{Jingkang Yang}, \bibinfo{person}{Chen~Change Loy}, {and} \bibinfo{person}{Ziwei Liu}.} \bibinfo{year}{2022}\natexlab{b}.
\newblock \showarticletitle{Learning to prompt for vision-language models}.
\newblock \bibinfo{journal}{\emph{International Journal of Computer Vision}} (\bibinfo{year}{2022}).
\newblock


\bibitem[Zhou et~al\mbox{.}(2022c)]%
        {zhou2022learning}
\bibfield{author}{\bibinfo{person}{Kaiyang Zhou}, \bibinfo{person}{Jingkang Yang}, \bibinfo{person}{Chen~Change Loy}, {and} \bibinfo{person}{Ziwei Liu}.} \bibinfo{year}{2022}\natexlab{c}.
\newblock \showarticletitle{Learning to prompt for vision-language models}.
\newblock \bibinfo{journal}{\emph{International Journal of Computer Vision}} (\bibinfo{year}{2022}).
\newblock


\bibitem[Zhu et~al\mbox{.}(2024a)]%
        {zhu2024relief}
\bibfield{author}{\bibinfo{person}{Jiapeng Zhu}, \bibinfo{person}{Zichen Ding}, \bibinfo{person}{Jianxiang Yu}, \bibinfo{person}{Jiaqi Tan}, \bibinfo{person}{Xiang Li}, {and} \bibinfo{person}{Weining Qian}.} \bibinfo{year}{2024}\natexlab{a}.
\newblock \showarticletitle{Relief: Reinforcement learning empowered graph feature prompt tuning}.
\newblock \bibinfo{journal}{\emph{arXiv preprint arXiv:2408.03195}} (\bibinfo{year}{2024}).
\newblock


\bibitem[Zhu et~al\mbox{.}(2024d)]%
        {zhu2024pitfalls}
\bibfield{author}{\bibinfo{person}{Jing Zhu}, \bibinfo{person}{Yuhang Zhou}, \bibinfo{person}{Vassilis~N Ioannidis}, \bibinfo{person}{Shengyi Qian}, \bibinfo{person}{Wei Ai}, \bibinfo{person}{Xiang Song}, {and} \bibinfo{person}{Danai Koutra}.} \bibinfo{year}{2024}\natexlab{d}.
\newblock \showarticletitle{Pitfalls in link prediction with graph neural networks: Understanding the impact of target-link inclusion \& better practices}. In \bibinfo{booktitle}{\emph{Proceedings of the 17th ACM International Conference on Web Search and Data Mining}}.
\newblock


\bibitem[Zhu et~al\mbox{.}(2020a)]%
        {zhu2020self}
\bibfield{author}{\bibinfo{person}{Qikui Zhu}, \bibinfo{person}{Bo Du}, {and} \bibinfo{person}{Pingkun Yan}.} \bibinfo{year}{2020}\natexlab{a}.
\newblock \showarticletitle{Self-supervised training of graph convolutional networks}.
\newblock \bibinfo{journal}{\emph{arXiv preprint arXiv:2006.02380}} (\bibinfo{year}{2020}).
\newblock


\bibitem[Zhu et~al\mbox{.}(2021b)]%
        {zhu2021transfer}
\bibfield{author}{\bibinfo{person}{Qi Zhu}, \bibinfo{person}{Carl Yang}, \bibinfo{person}{Yidan Xu}, \bibinfo{person}{Haonan Wang}, \bibinfo{person}{Chao Zhang}, {and} \bibinfo{person}{Jiawei Han}.} \bibinfo{year}{2021}\natexlab{b}.
\newblock \showarticletitle{Transfer learning of graph neural networks with ego-graph information maximization}.
\newblock \bibinfo{journal}{\emph{Advances in Neural Information Processing Systems}}  \bibinfo{volume}{34} (\bibinfo{year}{2021}), \bibinfo{pages}{1766--1779}.
\newblock


\bibitem[Zhu et~al\mbox{.}(2023)]%
        {zhu2023sgl}
\bibfield{author}{\bibinfo{person}{Yun Zhu}, \bibinfo{person}{Jianhao Guo}, {and} \bibinfo{person}{Siliang Tang}.} \bibinfo{year}{2023}\natexlab{}.
\newblock \showarticletitle{Sgl-pt: A strong graph learner with graph prompt tuning}.
\newblock \bibinfo{journal}{\emph{arXiv preprint arXiv:2302.12449}} (\bibinfo{year}{2023}).
\newblock


\bibitem[Zhu et~al\mbox{.}(2024b)]%
        {zhu2024collaborative}
\bibfield{author}{\bibinfo{person}{Yaochen Zhu}, \bibinfo{person}{Liang Wu}, \bibinfo{person}{Qi Guo}, \bibinfo{person}{Liangjie Hong}, {and} \bibinfo{person}{Jundong Li}.} \bibinfo{year}{2024}\natexlab{b}.
\newblock \showarticletitle{Collaborative large language model for recommender systems}. In \bibinfo{booktitle}{\emph{WWW}}. \bibinfo{pages}{3162--3172}.
\newblock


\bibitem[Zhu et~al\mbox{.}(2024c)]%
        {zhu2024understanding}
\bibfield{author}{\bibinfo{person}{Yaochen Zhu}, \bibinfo{person}{Liang Wu}, \bibinfo{person}{Binchi Zhang}, \bibinfo{person}{Song Wang}, \bibinfo{person}{Qi Guo}, \bibinfo{person}{Liangjie Hong}, \bibinfo{person}{Luke Simon}, {and} \bibinfo{person}{Jundong Li}.} \bibinfo{year}{2024}\natexlab{c}.
\newblock \showarticletitle{Understanding and Modeling Job Marketplace with Pretrained Language Models}. In \bibinfo{booktitle}{\emph{CIKM}}. \bibinfo{pages}{5143--5150}.
\newblock


\bibitem[Zhu et~al\mbox{.}(2020b)]%
        {zhu2020deep}
\bibfield{author}{\bibinfo{person}{Yanqiao Zhu}, \bibinfo{person}{Yichen Xu}, \bibinfo{person}{Feng Yu}, \bibinfo{person}{Qiang Liu}, \bibinfo{person}{Shu Wu}, {and} \bibinfo{person}{Liang Wang}.} \bibinfo{year}{2020}\natexlab{b}.
\newblock \showarticletitle{Deep graph contrastive representation learning}.
\newblock \bibinfo{journal}{\emph{arXiv preprint arXiv:2006.04131}} (\bibinfo{year}{2020}).
\newblock


\bibitem[Zhu et~al\mbox{.}(2021a)]%
        {zhu2021graph}
\bibfield{author}{\bibinfo{person}{Yanqiao Zhu}, \bibinfo{person}{Yichen Xu}, \bibinfo{person}{Feng Yu}, \bibinfo{person}{Qiang Liu}, \bibinfo{person}{Shu Wu}, {and} \bibinfo{person}{Liang Wang}.} \bibinfo{year}{2021}\natexlab{a}.
\newblock \showarticletitle{Graph contrastive learning with adaptive augmentation}. In \bibinfo{booktitle}{\emph{Proceedings of the web conference 2021}}. \bibinfo{pages}{2069--2080}.
\newblock


\bibitem[Zhuang et~al\mbox{.}(2023)]%
        {zhuang2024imold}
\bibfield{author}{\bibinfo{person}{Xiang Zhuang}, \bibinfo{person}{Qiang Zhang}, \bibinfo{person}{Keyan Ding}, \bibinfo{person}{Yatao Bian}, \bibinfo{person}{Xiao Wang}, \bibinfo{person}{Jingsong Lv}, \bibinfo{person}{Hongyang Chen}, {and} \bibinfo{person}{Huajun Chen}.} \bibinfo{year}{2023}\natexlab{}.
\newblock \showarticletitle{Learning invariant molecular representation in latent discrete space}.
\newblock \bibinfo{journal}{\emph{Advances in Neural Information Processing Systems}} (\bibinfo{year}{2023}).
\newblock


\bibitem[Zi et~al\mbox{.}(2024)]%
        {zi2024prog}
\bibfield{author}{\bibinfo{person}{Chenyi Zi}, \bibinfo{person}{Haihong Zhao}, \bibinfo{person}{Xiangguo Sun}, \bibinfo{person}{Yiqing Lin}, \bibinfo{person}{Hong Cheng}, {and} \bibinfo{person}{Jia Li}.} \bibinfo{year}{2024}\natexlab{}.
\newblock \showarticletitle{ProG: A Graph Prompt Learning Benchmark}.
\newblock \bibinfo{journal}{\emph{arXiv preprint arXiv:2406.05346}} (\bibinfo{year}{2024}).
\newblock


\bibitem[Z{\"u}gner and G{\"u}nnemann(2020)]%
        {zugner2020certifiable}
\bibfield{author}{\bibinfo{person}{Daniel Z{\"u}gner} {and} \bibinfo{person}{Stephan G{\"u}nnemann}.} \bibinfo{year}{2020}\natexlab{}.
\newblock \showarticletitle{Certifiable robustness of graph convolutional networks under structure perturbations}. In \bibinfo{booktitle}{\emph{Proceedings of the 26th ACM SIGKDD international conference on knowledge discovery \& data mining}}. \bibinfo{pages}{1656--1665}.
\newblock


\end{thebibliography}

\end{document}